\documentclass{article}

\usepackage{arxiv}
\usepackage[utf8]{inputenc} 
\usepackage[T1]{fontenc}    
\usepackage{url}            
\usepackage{booktabs}       
\usepackage{amsfonts}       
\usepackage{nicefrac}       
\usepackage{microtype}      
\usepackage{lipsum}		
\usepackage{graphicx}
\usepackage{doi}
\usepackage{upgreek,amsmath,bm}   
\usepackage[numbers]{natbib}
\usepackage{amssymb}
\usepackage{tabularx}
\usepackage{multirow,array,ragged2e,arydshln} 
\usepackage{float}
\usepackage{enumitem}

\newcommand\blfootnote[1]{%
  \begingroup
  \renewcommand\thefootnote{}\footnote{#1}%
  \addtocounter{footnote}{-1}%
  \endgroup
}

\usepackage{hyperref}
\hypersetup{
    colorlinks=true,
    linkcolor=blue,
    filecolor=magenta,      
    urlcolor=black,
    pdftitle={Overleaf Example},
    pdfpagemode=FullScreen,
    citecolor=blue,
    }

\title{Federated Learning with Uncertainty-Based Client Clustering for Fleet-Wide Fault Diagnosis}
\date{}

\author{
    \href{https://orcid.org/0000-0002-3557-531X}{\includegraphics[scale=0.06]{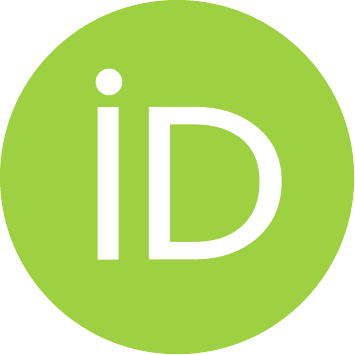}\hspace{1mm}Hao~Lu} \\
	Department of Electrical Engineering\\
	Iowa State University\\
	Ames, IA 50011\\
	\texttt{hlu1@iastate.edu} \\
	\And
	\href{https://orcid.org/0000-0001-5443-3678}{\includegraphics[scale=0.06]{FIGURES/orcid.pdf}\hspace{1mm}Adam~Thelen} \\
	Department of Mechanical Engineering\\
	Iowa State University\\
	Ames, IA 50011\\
	\texttt{acthelen@iastate.edu} \\
	\And
	\href{https://orcid.org/0000-0002-9546-1488}{\includegraphics[scale=0.06]{FIGURES/orcid.pdf}\hspace{1mm}Olga~Fink} \\
	Intelligent Maintenance and Operations Systems\\
	EPFL\\
	Lausanne, Switzerland 12309\\
	\texttt{olga.fink@epfl.ch} \\
	\And
	\href{https://orcid.org/0000-0001-9228-7675}{\includegraphics[scale=0.06]{FIGURES/orcid.pdf}\hspace{1mm}Chao~Hu$^{\dagger}$} \\
	Department of Mechanical Engineering\\
	University of Connecticut \\
	Storrs, CT 01776\\
	\texttt{chao.hu@uconn.edu} \\
        \And
	\href{https://orcid.org/0000-0002-0601-9664}{\includegraphics[scale=0.06]{FIGURES/orcid.pdf}\hspace{1mm}Simon~Laflamme} \\
	Department of Civil, Environmental, and Construction Engineering\\
	Iowa State University\\
	Ames, IA 50011\\
	\texttt{laflamme@iastate.edu} \\
 \\
 \normalsize $^{\dagger}$Indicates corresponding author. Email: \href{mailto:chao.hu@uconn.edu}{\textcolor{cyan}{chao.hu@uconn.edu}}.
}

\renewcommand{\undertitle}{}

\hypersetup{
pdftitle={Clustering-Based Distance-Aware Federated Learning for Bearing Diagnosis},
pdfsubject={Federated learning for fault diagnosis},
pdfauthor={Hao~Lu, Adam~Thelen, Olga~Fink, Chao~Hum,Simon~Laflamme},
pdfkeywords={federated learning, fault diagnosis, deep learning, clustering, industrial machinery},
}

\newcommand*\TableOne{
\begin{table}[h!]
\caption{Summary of selected bearing datasets}
\label{table:Dataset}
\centering
\resizebox{\textwidth}{!}{%
\begin{tabular}{cccccccc}
\hline
Name &
  Bearing health conditions &
  \begin{tabular}[c]{@{}c@{}}Cause of \\ bearing fault\end{tabular} &
  \begin{tabular}[c]{@{}c@{}}Sampling \\ frequency\end{tabular} &
  \multicolumn{4}{c}{Notes} \\ \hline
\multirow{7}{*}{CWRU} &
   &
   &
   &
  \begin{tabular}[c]{@{}c@{}}Working condition \\ ID\end{tabular} &
  \begin{tabular}[c]{@{}c@{}}Shaft \\ speed\end{tabular} &
  \multicolumn{2}{c}{\begin{tabular}[c]{@{}c@{}}Bearing fault \\ diameter\end{tabular}} \\ \cline{5-8} 
 &
  \multirow{6}{*}{\begin{tabular}[c]{@{}c@{}}Healthy, Inner race fault,\\ Outer race fault\end{tabular}} &
  \multirow{6}{*}{\begin{tabular}[c]{@{}c@{}}Electric \\ engraver\end{tabular}} &
  \multirow{6}{*}{12 kHz} &
  1 &
  1797 rpm &
  \multicolumn{2}{c}{0.007"} \\
 &  &  &  & 2 & 1797 rpm & \multicolumn{2}{c}{0.014"} \\
 &  &  &  & 3 & 1772 rpm & \multicolumn{2}{c}{0.007"} \\
 &  &  &  & 4 & 1772 rpm & \multicolumn{2}{c}{0.014"} \\
 &  &  &  & 5 & 1730 rpm & \multicolumn{2}{c}{0.007"} \\
 &  &  &  & 6 & 1730 rpm & \multicolumn{2}{c}{0.014"} \\ \hline
\multirow{5}{*}{PU} &
   &
   &
   &
  \begin{tabular}[c]{@{}c@{}}Working condition \\ ID\end{tabular} &
  \begin{tabular}[c]{@{}c@{}}Shaft \\ speed\end{tabular} &
  \begin{tabular}[c]{@{}c@{}}Load \\ torque\end{tabular} &
  \begin{tabular}[c]{@{}c@{}}Radial \\ force\end{tabular} \\ \cline{5-8} 
 &
  \multirow{4}{*}{\begin{tabular}[c]{@{}c@{}}Healthy, Inner race fault, \\ Outer race fault\end{tabular}} &
  \multirow{4}{*}{\begin{tabular}[c]{@{}c@{}}Accelerated \\ degradation\end{tabular}} &
  \multirow{4}{*}{64 kHz} &
  1 &
  1500 rpm &
  0.7 Nm &
  1000 N \\
 &  &  &  & 2 & 900 rpm  & 0.7 Nm       & 1000 N      \\
 &  &  &  & 3 & 1500 rpm & 0.1 Nm       & 1000 N      \\
 &  &  &  & 4 & 1500 rpm & 0.7 Nm       & 400 N       \\ \hline
\multirow{5}{*}{ISU} &
   &
   &
   &
  \begin{tabular}[c]{@{}c@{}}Working condition \\ ID\end{tabular} &
  \begin{tabular}[c]{@{}c@{}}Shaft \\ speed\end{tabular} &
  \multicolumn{2}{c}{\begin{tabular}[c]{@{}c@{}}Radial \\ force\end{tabular}} \\ \cline{5-8} 
 &
  \multirow{4}{*}{\begin{tabular}[c]{@{}c@{}}Healthy, Inner race fault, \\ Outer race fault, \\ Combination of bearing faults\end{tabular}} &
  \multirow{4}{*}{\begin{tabular}[c]{@{}c@{}}Electric \\ engraver\end{tabular}} &
  \multirow{4}{*}{25.6 kHz} &
  1 &
  2100 rpm &
  \multicolumn{2}{c}{0} \\
 &  &  &  & 2 & 2100 rpm & \multicolumn{2}{c}{25 N}   \\
 &  &  &  & 3 & 1500 rpm & \multicolumn{2}{c}{0}      \\
 &  &  &  & 4 & 1500 rpm & \multicolumn{2}{c}{25 N}   \\ \hline
\end{tabular}%
}
\end{table}
}

\newcommand*\TableTwo{

\begin{table}[h!]
\caption{Design of each client dataset for scenarios 1 and 2}
\label{table:Exp_design}
\centering
\begin{tabular}{ccccccccccccccc}
\hline
Dataset &
   &
   &
  \multicolumn{12}{c}{Number of samples} \\ \hline
\multirow{13}{*}{CWRU} &
  Client ID &
  Operating condition &
  \multicolumn{4}{c}{Healthy} &
  \multicolumn{4}{c}{\begin{tabular}[c]{@{}c@{}}Inner race \\ fault\end{tabular}} &
  \multicolumn{4}{c}{\begin{tabular}[c]{@{}c@{}}Outer race\\ fault\end{tabular}} \\ \cline{2-15} 
 &
  1 &
  \multirow{2}{*}{\begin{tabular}[c]{@{}c@{}}Shaft speed = 1797 rpm\\ Fault diameter = 0.007''\end{tabular}} &
  \multicolumn{4}{c}{80} &
  \multicolumn{4}{c}{80} &
  \multicolumn{4}{c}{0} \\
 &
  2 &
   &
  \multicolumn{4}{c}{80} &
  \multicolumn{4}{c}{0} &
  \multicolumn{4}{c}{80} \\ \cline{2-15} 
 &
  3 &
  \multirow{2}{*}{\begin{tabular}[c]{@{}c@{}}Shaft speed = 1797 rpm\\ Fault diameter = 0.014''\end{tabular}} &
  \multicolumn{4}{c}{80} &
  \multicolumn{4}{c}{80} &
  \multicolumn{4}{c}{0} \\
 &
  4 &
   &
  \multicolumn{4}{c}{80} &
  \multicolumn{4}{c}{0} &
  \multicolumn{4}{c}{80} \\ \cline{2-15} 
 &
  5 &
  \multirow{2}{*}{\begin{tabular}[c]{@{}c@{}}Shaft speed = 1772 rpm\\ Fault diameter = 0.007''\end{tabular}} &
  \multicolumn{4}{c}{80} &
  \multicolumn{4}{c}{80} &
  \multicolumn{4}{c}{0} \\
 &
  6 &
   &
  \multicolumn{4}{c}{80} &
  \multicolumn{4}{c}{0} &
  \multicolumn{4}{c}{80} \\ \cline{2-15} 
 &
  7 &
  \multirow{2}{*}{\begin{tabular}[c]{@{}c@{}}Shaft speed = 1772 rpm\\ Fault diameter = 0.014''\end{tabular}} &
  \multicolumn{4}{c}{80} &
  \multicolumn{4}{c}{80} &
  \multicolumn{4}{c}{0} \\
 &
  8 &
   &
  \multicolumn{4}{c}{80} &
  \multicolumn{4}{c}{0} &
  \multicolumn{4}{c}{80} \\ \cline{2-15} 
 &
  9 &
  \multirow{2}{*}{\begin{tabular}[c]{@{}c@{}}Shaft speed = 1730 rpm\\ Fault diameter = 0.007''\end{tabular}} &
  \multicolumn{4}{c}{80} &
  \multicolumn{4}{c}{80} &
  \multicolumn{4}{c}{0} \\
 &
  10 &
   &
  \multicolumn{4}{c}{80} &
  \multicolumn{4}{c}{0} &
  \multicolumn{4}{c}{80} \\ \cline{2-15} 
 &
  11 &
  \multirow{2}{*}{\begin{tabular}[c]{@{}c@{}}Shaft speed = 1730 rpm\\ Fault diameter = 0.014''\end{tabular}} &
  \multicolumn{4}{c}{80} &
  \multicolumn{4}{c}{80} &
  \multicolumn{4}{c}{0} \\
 &
  12 &
   &
  \multicolumn{4}{c}{80} &
  \multicolumn{4}{c}{0} &
  \multicolumn{4}{c}{80} \\ \hline
\multirow{13}{*}{PU} &
  Client ID &
  Operating condition &
  \multicolumn{4}{c}{Healthy} &
  \multicolumn{4}{c}{\begin{tabular}[c]{@{}c@{}}Inner race\\ fault\end{tabular}} &
  \multicolumn{4}{c}{\begin{tabular}[c]{@{}c@{}}Outer race\\ fault\end{tabular}} \\ \cline{2-15} 
 &
  1 &
  \multirow{3}{*}{\begin{tabular}[c]{@{}c@{}}Shaft speed = 1500 rpm\\ Load torque = 0.7 Nm\\ Radial force = 1000 N\end{tabular}} &
  \multicolumn{4}{c}{1200} &
  \multicolumn{4}{c}{1200} &
  \multicolumn{4}{c}{1200} \\
 &
  2 &
   &
  \multicolumn{4}{c}{1200} &
  \multicolumn{4}{c}{1200} &
  \multicolumn{4}{c}{0} \\
 &
  3 &
   &
  \multicolumn{4}{c}{1200} &
  \multicolumn{4}{c}{0} &
  \multicolumn{4}{c}{1200} \\ \cline{2-15} 
 &
  4 &
  \multirow{3}{*}{\begin{tabular}[c]{@{}c@{}}Shaft speed = 900 rpm\\ Load torque = 0.7 Nm\\ Radial force = 1000 N\end{tabular}} &
  \multicolumn{4}{c}{1200} &
  \multicolumn{4}{c}{1200} &
  \multicolumn{4}{c}{1200} \\
 &
  5 &
   &
  \multicolumn{4}{c}{1200} &
  \multicolumn{4}{c}{1200} &
  \multicolumn{4}{c}{0} \\
 &
  6 &
   &
  \multicolumn{4}{c}{1200} &
  \multicolumn{4}{c}{0} &
  \multicolumn{4}{c}{1200} \\ \cline{2-15} 
 &
  7 &
  \multirow{3}{*}{\begin{tabular}[c]{@{}c@{}}Shaft speed = 1500 rpm\\ Load torque = 0.1 Nm\\ Radial force = 1000 N\end{tabular}} &
  \multicolumn{4}{c}{1200} &
  \multicolumn{4}{c}{1200} &
  \multicolumn{4}{c}{1200} \\
 &
  8 &
   &
  \multicolumn{4}{c}{1200} &
  \multicolumn{4}{c}{1200} &
  \multicolumn{4}{c}{0} \\
 &
  9 &
   &
  \multicolumn{4}{c}{1200} &
  \multicolumn{4}{c}{0} &
  \multicolumn{4}{c}{1200} \\ \cline{2-15} 
 &
  10 &
  \multirow{3}{*}{\begin{tabular}[c]{@{}c@{}}Shaft speed = 1500 rpm\\ Load torque = 0.7 Nm\\ Radial force = 400N\end{tabular}} &
  \multicolumn{4}{c}{1200} &
  \multicolumn{4}{c}{1200} &
  \multicolumn{4}{c}{1200} \\
 &
  11 &
   &
  \multicolumn{4}{c}{1200} &
  \multicolumn{4}{c}{1200} &
  \multicolumn{4}{c}{0} \\
 &
  12 &
   &
  \multicolumn{4}{c}{1200} &
  \multicolumn{4}{c}{0} &
  \multicolumn{4}{c}{1200} \\ \hline
\multirow{13}{*}{ISU} &
  Client ID &
  Operating condition &
  \multicolumn{3}{c}{Healthy} &
  \multicolumn{3}{c}{\begin{tabular}[c]{@{}c@{}}Inner race\\ fault\end{tabular}} &
  \multicolumn{3}{c}{\begin{tabular}[c]{@{}c@{}}Outer race\\ fault\end{tabular}} &
  \multicolumn{3}{c}{\begin{tabular}[c]{@{}c@{}}Combination \\ of faults\end{tabular}} \\ \cline{2-15} 
 &
  1 &
  \multirow{3}{*}{\begin{tabular}[c]{@{}c@{}}Shaft speed = 2100 rpm\\ Radial force = 0\end{tabular}} &
  \multicolumn{3}{c}{2000} &
  \multicolumn{3}{c}{2000} &
  \multicolumn{3}{c}{0} &
  \multicolumn{3}{c}{0} \\
 &
  2 &
   &
  \multicolumn{3}{c}{2000} &
  \multicolumn{3}{c}{0} &
  \multicolumn{3}{c}{2000} &
  \multicolumn{3}{c}{0} \\
 &
  3 &
   &
  \multicolumn{3}{c}{2000} &
  \multicolumn{3}{c}{0} &
  \multicolumn{3}{c}{0} &
  \multicolumn{3}{c}{2000} \\ \cline{2-15} 
 &
  4 &
  \multirow{3}{*}{\begin{tabular}[c]{@{}c@{}}Shaft speed = 2100 rpm\\ Radial force = 25 N\end{tabular}} &
  \multicolumn{3}{c}{2000} &
  \multicolumn{3}{c}{2000} &
  \multicolumn{3}{c}{0} &
  \multicolumn{3}{c}{0} \\
 &
  5 &
   &
  \multicolumn{3}{c}{2000} &
  \multicolumn{3}{c}{0} &
  \multicolumn{3}{c}{2000} &
  \multicolumn{3}{c}{0} \\
 &
  6 &
   &
  \multicolumn{3}{c}{2000} &
  \multicolumn{3}{c}{0} &
  \multicolumn{3}{c}{0} &
  \multicolumn{3}{c}{2000} \\ \cline{2-15} 
 &
  7 &
  \multirow{3}{*}{\begin{tabular}[c]{@{}c@{}}Shaft speed = 1500 rpm\\ Radial force = 0\end{tabular}} &
  \multicolumn{3}{c}{2000} &
  \multicolumn{3}{c}{2000} &
  \multicolumn{3}{c}{0} &
  \multicolumn{3}{c}{0} \\
 &
  8 &
   &
  \multicolumn{3}{c}{2000} &
  \multicolumn{3}{c}{0} &
  \multicolumn{3}{c}{2000} &
  \multicolumn{3}{c}{0} \\
 &
  9 &
   &
  \multicolumn{3}{c}{2000} &
  \multicolumn{3}{c}{0} &
  \multicolumn{3}{c}{0} &
  \multicolumn{3}{c}{2000} \\ \cline{2-15} 
 &
  10 &
  \multirow{3}{*}{\begin{tabular}[c]{@{}c@{}}Shaft speed = 1500 rpm\\ Radial force = 25 N\end{tabular}} &
  \multicolumn{3}{c}{2000} &
  \multicolumn{3}{c}{2000} &
  \multicolumn{3}{c}{0} &
  \multicolumn{3}{c}{0} \\
 &
  11 &
   &
  \multicolumn{3}{c}{2000} &
  \multicolumn{3}{c}{0} &
  \multicolumn{3}{c}{2000} &
  \multicolumn{3}{c}{0} \\
 &
  12 &
   &
  \multicolumn{3}{c}{2000} &
  \multicolumn{3}{c}{0} &
  \multicolumn{3}{c}{0} &
  \multicolumn{3}{c}{2000} \\ \hline
 &
   &
   &
   &
   &
   &
   &
   &
   &
   &
   &
   &
   &
   &
  
\end{tabular}

\end{table}

}

\newcommand*\TableThree{
\begin{table}[h!]
\caption{Summarization of average test accuracy for each case study}
\label{table:Average_acc}
\centering
\begin{tabular}{llllllllll}
\hline
\multicolumn{1}{c}{\multirow{2}{*}{Methods}} & \multicolumn{3}{c}{Scenario 1} & \multicolumn{3}{c}{Scenario 2} & \multicolumn{3}{c}{Scenario 3} \\ \cline{2-10} 
\multicolumn{1}{c}{}                         & CWRU     & PU       & ISU      & CWRU     & PU       & ISU      & CWRU     & PU       & ISU      \\ \hline
Local training                               & 99.58    & 99.93    & 100.00   & 75.69    & 77.72    & 50.11    & 70.56    & 71.33    & 45.76    \\
FedAvg                                       & 99.37    & 98.31    & 98.80    & 99.16    & 98.14    & 98.20    & 89.17    & 97.67    & 92.80    \\
FedCos                                       & 99.79    & 99.47    & 99.92    & 99.72    & 80.06    & 50.00    & 72.63    & 80.39    & 49.98    \\
FedSNGP                                      & 99.58    & 99.90    & 100.00   & 99.44    & 99.92    & 100.00   & 95.56    & 99.81    & 99.66    \\ \hline
\end{tabular}
\end{table}
}

\usepackage{multirow}
\usepackage{graphicx}
\newcommand*\Appendtestdataset{
\begin{table}[H]
\label{table:Design_test}
\centering
\resizebox{\textwidth}{!}{%
\begin{tabular}{@{}ccccccccccccccccccccccccccc@{}}
\toprule
Dataset &
   &
   &
  \multicolumn{12}{c}{Scenario 1 test dataset} &
  \multicolumn{12}{c}{Scenario 2 and 3 test dataset} \\ \midrule
\multirow{13}{*}{CWRU} &
  Client ID &
  Operating condition &
  \multicolumn{4}{c}{Healthy} &
  \multicolumn{4}{c}{\begin{tabular}[c]{@{}c@{}}Inner race \\ fault\end{tabular}} &
  \multicolumn{4}{c}{\begin{tabular}[c]{@{}c@{}}Outer race\\ fault\end{tabular}} &
  \multicolumn{4}{c}{Healthy} &
  \multicolumn{4}{c}{\begin{tabular}[c]{@{}c@{}}Inner race \\ fault\end{tabular}} &
  \multicolumn{4}{c}{\begin{tabular}[c]{@{}c@{}}Outer race\\ fault\end{tabular}} \\
 &
  1 &
  \multirow{2}{*}{\begin{tabular}[c]{@{}c@{}}Shaft speed = 1797 rpm\\ Fault diameter = 0.007''\end{tabular}} &
  \multicolumn{4}{c}{20} &
  \multicolumn{4}{c}{20} &
  \multicolumn{4}{c}{0} &
  \multicolumn{4}{c}{20} &
  \multicolumn{4}{c}{20} &
  \multicolumn{4}{c}{20} \\
 &
  2 &
   &
  \multicolumn{4}{c}{20} &
  \multicolumn{4}{c}{0} &
  \multicolumn{4}{c}{20} &
  \multicolumn{4}{c}{20} &
  \multicolumn{4}{c}{20} &
  \multicolumn{4}{c}{20} \\
 &
  3 &
  \multirow{2}{*}{\begin{tabular}[c]{@{}c@{}}Shaft speed = 1797 rpm\\ Fault diameter = 0.014''\end{tabular}} &
  \multicolumn{4}{c}{20} &
  \multicolumn{4}{c}{20} &
  \multicolumn{4}{c}{0} &
  \multicolumn{4}{c}{20} &
  \multicolumn{4}{c}{20} &
  \multicolumn{4}{c}{20} \\
 &
  4 &
   &
  \multicolumn{4}{c}{20} &
  \multicolumn{4}{c}{0} &
  \multicolumn{4}{c}{20} &
  \multicolumn{4}{c}{20} &
  \multicolumn{4}{c}{20} &
  \multicolumn{4}{c}{20} \\
 &
  5 &
  \multirow{2}{*}{\begin{tabular}[c]{@{}c@{}}Shaft speed = 1772 rpm\\ Fault diameter = 0.007''\end{tabular}} &
  \multicolumn{4}{c}{20} &
  \multicolumn{4}{c}{20} &
  \multicolumn{4}{c}{0} &
  \multicolumn{4}{c}{20} &
  \multicolumn{4}{c}{20} &
  \multicolumn{4}{c}{20} \\
 &
  6 &
   &
  \multicolumn{4}{c}{20} &
  \multicolumn{4}{c}{0} &
  \multicolumn{4}{c}{20} &
  \multicolumn{4}{c}{20} &
  \multicolumn{4}{c}{20} &
  \multicolumn{4}{c}{20} \\
 &
  7 &
  \multirow{2}{*}{\begin{tabular}[c]{@{}c@{}}Shaft speed = 1772 rpm\\ Fault diameter = 0.014''\end{tabular}} &
  \multicolumn{4}{c}{20} &
  \multicolumn{4}{c}{20} &
  \multicolumn{4}{c}{0} &
  \multicolumn{4}{c}{20} &
  \multicolumn{4}{c}{20} &
  \multicolumn{4}{c}{20} \\
 &
  8 &
   &
  \multicolumn{4}{c}{20} &
  \multicolumn{4}{c}{0} &
  \multicolumn{4}{c}{20} &
  \multicolumn{4}{c}{20} &
  \multicolumn{4}{c}{20} &
  \multicolumn{4}{c}{20} \\
 &
  9 &
  \multirow{2}{*}{\begin{tabular}[c]{@{}c@{}}Shaft speed = 1730 rpm\\ Fault diameter = 0.007''\end{tabular}} &
  \multicolumn{4}{c}{20} &
  \multicolumn{4}{c}{20} &
  \multicolumn{4}{c}{0} &
  \multicolumn{4}{c}{20} &
  \multicolumn{4}{c}{20} &
  \multicolumn{4}{c}{20} \\
 &
  10 &
   &
  \multicolumn{4}{c}{20} &
  \multicolumn{4}{c}{0} &
  \multicolumn{4}{c}{20} &
  \multicolumn{4}{c}{20} &
  \multicolumn{4}{c}{20} &
  \multicolumn{4}{c}{20} \\
 &
  11 &
  \multirow{2}{*}{\begin{tabular}[c]{@{}c@{}}Shaft speed = 1730 rpm\\ Fault diameter = 0.014''\end{tabular}} &
  \multicolumn{4}{c}{20} &
  \multicolumn{4}{c}{80} &
  \multicolumn{4}{c}{0} &
  \multicolumn{4}{c}{20} &
  \multicolumn{4}{c}{20} &
  \multicolumn{4}{c}{20} \\
 &
  12 &
   &
  \multicolumn{4}{c}{20} &
  \multicolumn{4}{c}{0} &
  \multicolumn{4}{c}{20} &
  \multicolumn{4}{c}{20} &
  \multicolumn{4}{c}{20} &
  \multicolumn{4}{c}{20} \\ \midrule
\multirow{13}{*}{PU} &
  Client ID &
  Operating condition &
  \multicolumn{4}{c}{Healthy} &
  \multicolumn{4}{c}{\begin{tabular}[c]{@{}c@{}}Inner race\\ fault\end{tabular}} &
  \multicolumn{4}{c}{\begin{tabular}[c]{@{}c@{}}Outer race\\ fault\end{tabular}} &
  \multicolumn{4}{c}{Healthy} &
  \multicolumn{4}{c}{\begin{tabular}[c]{@{}c@{}}Inner race\\ fault\end{tabular}} &
  \multicolumn{4}{c}{\begin{tabular}[c]{@{}c@{}}Outer race\\ fault\end{tabular}} \\
 &
  1 &
  \multirow{3}{*}{\begin{tabular}[c]{@{}c@{}}Shaft speed = 1500 rpm\\ Load torque = 0.7 Nm\\ Radial force = 1000 N\end{tabular}} &
  \multicolumn{4}{c}{300} &
  \multicolumn{4}{c}{300} &
  \multicolumn{4}{c}{300} &
  \multicolumn{4}{c}{300} &
  \multicolumn{4}{c}{300} &
  \multicolumn{4}{c}{300} \\
 &
  2 &
   &
  \multicolumn{4}{c}{300} &
  \multicolumn{4}{c}{300} &
  \multicolumn{4}{c}{0} &
  \multicolumn{4}{c}{300} &
  \multicolumn{4}{c}{300} &
  \multicolumn{4}{c}{300} \\
 &
  3 &
   &
  \multicolumn{4}{c}{300} &
  \multicolumn{4}{c}{0} &
  \multicolumn{4}{c}{300} &
  \multicolumn{4}{c}{300} &
  \multicolumn{4}{c}{300} &
  \multicolumn{4}{c}{300} \\
 &
  4 &
  \multirow{3}{*}{\begin{tabular}[c]{@{}c@{}}Shaft speed = 900 rpm\\ Load torque = 0.7 Nm\\ Radial force = 1000 N\end{tabular}} &
  \multicolumn{4}{c}{300} &
  \multicolumn{4}{c}{300} &
  \multicolumn{4}{c}{300} &
  \multicolumn{4}{c}{300} &
  \multicolumn{4}{c}{300} &
  \multicolumn{4}{c}{300} \\
 &
  5 &
   &
  \multicolumn{4}{c}{300} &
  \multicolumn{4}{c}{300} &
  \multicolumn{4}{c}{0} &
  \multicolumn{4}{c}{300} &
  \multicolumn{4}{c}{300} &
  \multicolumn{4}{c}{300} \\
 &
  6 &
   &
  \multicolumn{4}{c}{300} &
  \multicolumn{4}{c}{0} &
  \multicolumn{4}{c}{300} &
  \multicolumn{4}{c}{300} &
  \multicolumn{4}{c}{300} &
  \multicolumn{4}{c}{300} \\
 &
  7 &
  \multirow{3}{*}{\begin{tabular}[c]{@{}c@{}}Shaft speed = 1500 rpm\\ Load torque = 0.1 Nm\\ Radial force = 1000 N\end{tabular}} &
  \multicolumn{4}{c}{300} &
  \multicolumn{4}{c}{300} &
  \multicolumn{4}{c}{300} &
  \multicolumn{4}{c}{300} &
  \multicolumn{4}{c}{300} &
  \multicolumn{4}{c}{300} \\
 &
  8 &
   &
  \multicolumn{4}{c}{300} &
  \multicolumn{4}{c}{300} &
  \multicolumn{4}{c}{0} &
  \multicolumn{4}{c}{300} &
  \multicolumn{4}{c}{300} &
  \multicolumn{4}{c}{300} \\
 &
  9 &
   &
  \multicolumn{4}{c}{300} &
  \multicolumn{4}{c}{0} &
  \multicolumn{4}{c}{300} &
  \multicolumn{4}{c}{300} &
  \multicolumn{4}{c}{300} &
  \multicolumn{4}{c}{300} \\
 &
  10 &
  \multirow{3}{*}{\begin{tabular}[c]{@{}c@{}}Shaft speed = 1500 rpm\\ Load torque = 0.7 Nm\\ Radial force = 400N\end{tabular}} &
  \multicolumn{4}{c}{300} &
  \multicolumn{4}{c}{300} &
  \multicolumn{4}{c}{300} &
  \multicolumn{4}{c}{300} &
  \multicolumn{4}{c}{300} &
  \multicolumn{4}{c}{300} \\
 &
  11 &
   &
  \multicolumn{4}{c}{300} &
  \multicolumn{4}{c}{300} &
  \multicolumn{4}{c}{0} &
  \multicolumn{4}{c}{300} &
  \multicolumn{4}{c}{300} &
  \multicolumn{4}{c}{300} \\
 &
  12 &
   &
  \multicolumn{4}{c}{300} &
  \multicolumn{4}{c}{0} &
  \multicolumn{4}{c}{300} &
  \multicolumn{4}{c}{300} &
  \multicolumn{4}{c}{300} &
  \multicolumn{4}{c}{300} \\ \midrule
\multirow{13}{*}{ISU} &
  Client ID &
  Operating condition &
  \multicolumn{3}{c}{Healthy} &
  \multicolumn{3}{c}{\begin{tabular}[c]{@{}c@{}}Inner race\\ fault\end{tabular}} &
  \multicolumn{3}{c}{\begin{tabular}[c]{@{}c@{}}Outer race\\ fault\end{tabular}} &
  \multicolumn{3}{c}{\begin{tabular}[c]{@{}c@{}}Combination \\ of faults\end{tabular}} &
  \multicolumn{3}{c}{Healthy} &
  \multicolumn{3}{c}{\begin{tabular}[c]{@{}c@{}}Inner race\\ fault\end{tabular}} &
  \multicolumn{3}{c}{\begin{tabular}[c]{@{}c@{}}Outer race\\ fault\end{tabular}} &
  \multicolumn{3}{c}{\begin{tabular}[c]{@{}c@{}}Combination \\ of faults\end{tabular}} \\
 &
  1 &
  \multirow{3}{*}{\begin{tabular}[c]{@{}c@{}}Shaft speed = 2100 rpm\\ Radial force = 0\end{tabular}} &
  \multicolumn{3}{c}{500} &
  \multicolumn{3}{c}{500} &
  \multicolumn{3}{c}{0} &
  \multicolumn{3}{c}{0} &
  \multicolumn{3}{c}{500} &
  \multicolumn{3}{c}{500} &
  \multicolumn{3}{c}{500} &
  \multicolumn{3}{c}{500} \\
 &
  2 &
   &
  \multicolumn{3}{c}{500} &
  \multicolumn{3}{c}{0} &
  \multicolumn{3}{c}{500} &
  \multicolumn{3}{c}{0} &
  \multicolumn{3}{c}{500} &
  \multicolumn{3}{c}{500} &
  \multicolumn{3}{c}{500} &
  \multicolumn{3}{c}{500} \\
 &
  3 &
   &
  \multicolumn{3}{c}{500} &
  \multicolumn{3}{c}{0} &
  \multicolumn{3}{c}{0} &
  \multicolumn{3}{c}{500} &
  \multicolumn{3}{c}{500} &
  \multicolumn{3}{c}{500} &
  \multicolumn{3}{c}{500} &
  \multicolumn{3}{c}{500} \\
 &
  4 &
  \multirow{3}{*}{\begin{tabular}[c]{@{}c@{}}Shaft speed = 2100 rpm\\ Radial force = 25 N\end{tabular}} &
  \multicolumn{3}{c}{500} &
  \multicolumn{3}{c}{500} &
  \multicolumn{3}{c}{0} &
  \multicolumn{3}{c}{0} &
  \multicolumn{3}{c}{500} &
  \multicolumn{3}{c}{500} &
  \multicolumn{3}{c}{500} &
  \multicolumn{3}{c}{500} \\
 &
  5 &
   &
  \multicolumn{3}{c}{500} &
  \multicolumn{3}{c}{0} &
  \multicolumn{3}{c}{500} &
  \multicolumn{3}{c}{0} &
  \multicolumn{3}{c}{500} &
  \multicolumn{3}{c}{500} &
  \multicolumn{3}{c}{500} &
  \multicolumn{3}{c}{500} \\
 &
  6 &
   &
  \multicolumn{3}{c}{500} &
  \multicolumn{3}{c}{0} &
  \multicolumn{3}{c}{0} &
  \multicolumn{3}{c}{500} &
  \multicolumn{3}{c}{500} &
  \multicolumn{3}{c}{500} &
  \multicolumn{3}{c}{500} &
  \multicolumn{3}{c}{500} \\
 &
  7 &
  \multirow{3}{*}{\begin{tabular}[c]{@{}c@{}}Shaft speed = 1500 rpm\\ Radial force = 0\end{tabular}} &
  \multicolumn{3}{c}{500} &
  \multicolumn{3}{c}{500} &
  \multicolumn{3}{c}{0} &
  \multicolumn{3}{c}{0} &
  \multicolumn{3}{c}{500} &
  \multicolumn{3}{c}{500} &
  \multicolumn{3}{c}{500} &
  \multicolumn{3}{c}{500} \\
 &
  8 &
   &
  \multicolumn{3}{c}{500} &
  \multicolumn{3}{c}{0} &
  \multicolumn{3}{c}{500} &
  \multicolumn{3}{c}{0} &
  \multicolumn{3}{c}{500} &
  \multicolumn{3}{c}{500} &
  \multicolumn{3}{c}{500} &
  \multicolumn{3}{c}{500} \\
 &
  9 &
   &
  \multicolumn{3}{c}{500} &
  \multicolumn{3}{c}{0} &
  \multicolumn{3}{c}{0} &
  \multicolumn{3}{c}{500} &
  \multicolumn{3}{c}{500} &
  \multicolumn{3}{c}{500} &
  \multicolumn{3}{c}{500} &
  \multicolumn{3}{c}{500} \\
 &
  10 &
  \multirow{3}{*}{\begin{tabular}[c]{@{}c@{}}Shaft speed = 1500 rpm\\ Radial force = 25 N\end{tabular}} &
  \multicolumn{3}{c}{500} &
  \multicolumn{3}{c}{500} &
  \multicolumn{3}{c}{0} &
  \multicolumn{3}{c}{0} &
  \multicolumn{3}{c}{500} &
  \multicolumn{3}{c}{500} &
  \multicolumn{3}{c}{500} &
  \multicolumn{3}{c}{500} \\
 &
  11 &
   &
  \multicolumn{3}{c}{500} &
  \multicolumn{3}{c}{0} &
  \multicolumn{3}{c}{500} &
  \multicolumn{3}{c}{0} &
  \multicolumn{3}{c}{500} &
  \multicolumn{3}{c}{500} &
  \multicolumn{3}{c}{500} &
  \multicolumn{3}{c}{500} \\
 &
  12 &
   &
  \multicolumn{3}{c}{500} &
  \multicolumn{3}{c}{0} &
  \multicolumn{3}{c}{0} &
  \multicolumn{3}{c}{500} &
  \multicolumn{3}{c}{500} &
  \multicolumn{3}{c}{500} &
  \multicolumn{3}{c}{500} &
  \multicolumn{3}{c}{500} \\ \midrule
 &
   &
   &
   &
   &
   &
   &
   &
   &
   &
   &
   &
   &
   &
   &
   &
   &
   &
   &
   &
   &
   &
   &
   &
   &
   &
  
\end{tabular}%
}
\end{table}
}

\newcommand*\Appendtrainingsc{
\begin{table}[H]
\centering
\label{table:Design_sc3train}
\resizebox{\textwidth}{!}{%
\begin{tabular}{cclllclllclllclllclllclllcllcllcllcll}
 &
  \multicolumn{12}{c}{CWRU dataset case study} &
  \multicolumn{12}{c}{PU dataset case study} &
  \multicolumn{12}{c}{ISU dataset case study} \\ \hline
Client ID &
  \multicolumn{4}{c}{Healthy} &
  \multicolumn{4}{c}{\begin{tabular}[c]{@{}c@{}}Inner race \\ fault\end{tabular}} &
  \multicolumn{4}{c}{\begin{tabular}[c]{@{}c@{}}Outer race\\ fault\end{tabular}} &
  \multicolumn{4}{c}{Healthy} &
  \multicolumn{4}{c}{\begin{tabular}[c]{@{}c@{}}Inner race\\ fault\end{tabular}} &
  \multicolumn{4}{c}{\begin{tabular}[c]{@{}c@{}}Outer race\\ fault\end{tabular}} &
  \multicolumn{3}{c}{Healthy} &
  \multicolumn{3}{c}{\begin{tabular}[c]{@{}c@{}}Inner race\\ fault\end{tabular}} &
  \multicolumn{3}{c}{\begin{tabular}[c]{@{}c@{}}Outer race\\ fault\end{tabular}} &
  \multicolumn{3}{c}{\begin{tabular}[c]{@{}c@{}}Combination \\ of faults\end{tabular}} \\
1 &
  \multicolumn{4}{c}{64} &
  \multicolumn{4}{c}{64} &
  \multicolumn{4}{c}{0} &
  \multicolumn{4}{c}{480} &
  \multicolumn{4}{c}{480} &
  \multicolumn{4}{c}{480} &
  \multicolumn{3}{c}{1200} &
  \multicolumn{3}{c}{1200} &
  \multicolumn{3}{c}{0} &
  \multicolumn{3}{c}{0} \\
2 &
  \multicolumn{4}{c}{64} &
  \multicolumn{4}{c}{0} &
  \multicolumn{4}{c}{64} &
  \multicolumn{4}{c}{480} &
  \multicolumn{4}{c}{480} &
  \multicolumn{4}{c}{0} &
  \multicolumn{3}{c}{1200} &
  \multicolumn{3}{c}{0} &
  \multicolumn{3}{c}{1200} &
  \multicolumn{3}{c}{0} \\
3 &
  \multicolumn{4}{c}{64} &
  \multicolumn{4}{c}{64} &
  \multicolumn{4}{c}{0} &
  \multicolumn{4}{c}{480} &
  \multicolumn{4}{c}{0} &
  \multicolumn{4}{c}{480} &
  \multicolumn{3}{c}{1200} &
  \multicolumn{3}{c}{0} &
  \multicolumn{3}{c}{0} &
  \multicolumn{3}{c}{1200} \\
4 &
  \multicolumn{4}{c}{56} &
  \multicolumn{4}{c}{0} &
  \multicolumn{4}{c}{56} &
  \multicolumn{4}{c}{1200} &
  \multicolumn{4}{c}{1200} &
  \multicolumn{4}{c}{1200} &
  \multicolumn{3}{c}{1800} &
  \multicolumn{3}{c}{1800} &
  \multicolumn{3}{c}{0} &
  \multicolumn{3}{c}{0} \\
5 &
  \multicolumn{4}{c}{56} &
  \multicolumn{4}{c}{56} &
  \multicolumn{4}{c}{0} &
  \multicolumn{4}{c}{1200} &
  \multicolumn{4}{c}{1200} &
  \multicolumn{4}{c}{0} &
  \multicolumn{3}{c}{1800} &
  \multicolumn{3}{c}{0} &
  \multicolumn{3}{c}{1800} &
  \multicolumn{3}{c}{0} \\
6 &
  \multicolumn{4}{c}{56} &
  \multicolumn{4}{c}{0} &
  \multicolumn{4}{c}{56} &
  \multicolumn{4}{c}{1200} &
  \multicolumn{4}{c}{0} &
  \multicolumn{4}{c}{1200} &
  \multicolumn{3}{c}{1800} &
  \multicolumn{3}{c}{0} &
  \multicolumn{3}{c}{0} &
  \multicolumn{3}{c}{1800} \\
7 &
  \multicolumn{4}{c}{64} &
  \multicolumn{4}{c}{64} &
  \multicolumn{4}{c}{0} &
  \multicolumn{4}{c}{600} &
  \multicolumn{4}{c}{600} &
  \multicolumn{4}{c}{600} &
  \multicolumn{3}{c}{800} &
  \multicolumn{3}{c}{800} &
  \multicolumn{3}{c}{0} &
  \multicolumn{3}{c}{0} \\
8 &
  \multicolumn{4}{c}{64} &
  \multicolumn{4}{c}{0} &
  \multicolumn{4}{c}{64} &
  \multicolumn{4}{c}{600} &
  \multicolumn{4}{c}{600} &
  \multicolumn{4}{c}{0} &
  \multicolumn{3}{c}{800} &
  \multicolumn{3}{c}{0} &
  \multicolumn{3}{c}{800} &
  \multicolumn{3}{c}{0} \\
9 &
  \multicolumn{4}{c}{64} &
  \multicolumn{4}{c}{64} &
  \multicolumn{4}{c}{0} &
  \multicolumn{4}{c}{600} &
  \multicolumn{4}{c}{0} &
  \multicolumn{4}{c}{600} &
  \multicolumn{3}{c}{800} &
  \multicolumn{3}{c}{0} &
  \multicolumn{3}{c}{0} &
  \multicolumn{3}{c}{800} \\
10 &
  \multicolumn{4}{c}{72} &
  \multicolumn{4}{c}{0} &
  \multicolumn{4}{c}{72} &
  \multicolumn{4}{c}{720} &
  \multicolumn{4}{c}{720} &
  \multicolumn{4}{c}{720} &
  \multicolumn{3}{c}{1600} &
  \multicolumn{3}{c}{1600} &
  \multicolumn{3}{c}{0} &
  \multicolumn{3}{c}{0} \\
11 &
  \multicolumn{4}{c}{72} &
  \multicolumn{4}{c}{72} &
  \multicolumn{4}{c}{0} &
  \multicolumn{4}{c}{720} &
  \multicolumn{4}{c}{720} &
  \multicolumn{4}{c}{0} &
  \multicolumn{3}{c}{1600} &
  \multicolumn{3}{c}{0} &
  \multicolumn{3}{c}{1600} &
  \multicolumn{3}{c}{0} \\
12 &
  \multicolumn{4}{c}{72} &
  \multicolumn{4}{c}{0} &
  \multicolumn{4}{c}{72} &
  \multicolumn{4}{c}{720} &
  \multicolumn{4}{c}{0} &
  \multicolumn{4}{c}{720} &
  \multicolumn{3}{c}{1600} &
  \multicolumn{3}{c}{0} &
  \multicolumn{3}{c}{0} &
  \multicolumn{3}{c}{1600} \\ \hline
\end{tabular}%
}
\end{table}
}

\newcommand*\Appendresults{
\begin{table}[H]
\centering
\label{table:All_results}
\resizebox{\textwidth}{!}{%
\begin{tabular}{ccccccccccccc}
\hline
\multicolumn{13}{c}{CWRU} \\ \hline
 & \multicolumn{4}{c}{Scenario 1} & \multicolumn{4}{c}{Scenario 2} & \multicolumn{4}{c}{Scenario 3} \\ \cline{2-13} 
 & Local training & FedAvg & FedCos & FedSNGP & Local training & FedAvg & FedCos & FedSNGP & Local training & FedAvg & FedCos & FedSNGP \\
1 & 100.00 & 100.00 & 100.00 & 100.00 & 81.67 & 100.00 & 100.00 & 100.00 & 78.33 & 93.33 & 71.67 & 96.67 \\
2 & 95.00 & 97.50 & 100.00 & 97.50 & 73.33 & 98.33 & 100.00 & 98.33 & 65.00 & 95.00 & 66.67 & 98.33 \\
3 & 100.00 & 100.00 & 100.00 & 100.00 & 75.00 & 100.00 & 100.00 & 100.00 & 75.00 & 93.33 & 66.67 & 96.67 \\
4 & 100.00 & 100.00 & 100.00 & 100.00 & 85.00 & 98.33 & 100.00 & 98.33 & 75.00 & 95.00 & 66.67 & 98.33 \\
5 & 100.00 & 100.00 & 100.00 & 100.00 & 76.67 & 100.00 & 100.00 & 100.00 & 73.33 & 93.33 & 100.00 & 100.00 \\
6 & 100.00 & 100.00 & 100.00 & 97.50 & 75.00 & 98.33 & 98.33 & 98.33 & 58.33 & 80.00 & 66.67 & 85.00 \\
7 & 100.00 & 100.00 & 100.00 & 100.00 & 71.67 & 100.00 & 100.00 & 100.00 & 58.33 & 93.33 & 100.00 & 100.00 \\
8 & 100.00 & 97.50 & 97.50 & 100.00 & 73.33 & 98.33 & 98.33 & 98.33 & 71.67 & 80.00 & 66.67 & 85.00 \\
9 & 100.00 & 100.00 & 100.00 & 100.00 & 71.67 & 100.00 & 100.00 & 100.00 & 70.00 & 90.00 & 66.67 & 98.33 \\
10 & 100.00 & 100.00 & 100.00 & 100.00 & 78.33 & 98.33 & 100.00 & 100.00 & 76.67 & 83.33 & 66.67 & 95.00 \\
11 & 100.00 & 100.00 & 100.00 & 100.00 & 76.67 & 100.00 & 100.00 & 100.00 & 68.33 & 90.00 & 66.67 & 98.33 \\
12 & 100.00 & 97.50 & 100.00 & 100.00 & 70.00 & 98.33 & 100.00 & 100.00 & 76.67 & 83.33 & 66.67 & 95.00 \\ \hline
\multicolumn{13}{c}{PU} \\ \hline
 & \multicolumn{4}{c}{Scenario 1} & \multicolumn{4}{c}{Scenario 2} & \multicolumn{4}{c}{Scenario 3} \\ \cline{2-13} 
 & Local training & FedAvg & FedCos & FedSNGP & Local training & FedAvg & FedCos & FedSNGP & Local training & FedAvg & FedCos & FedSNGP \\
1 & 99.89 & 99.78 & 100.00 & 100.00 & 99.89 & 99.78 & 100.00 & 100.00 & 100.00 & 97.56 & 99.56 & 99.89 \\
2 & 100.00 & 99.65 & 100.00 & 100.00 & 66.67 & 99.78 & 66.67 & 100.00 & 51.00 & 97.56 & 66.89 & 99.89 \\
3 & 100.00 & 100.00 & 100.00 & 100.00 & 66.67 & 99.78 & 66.56 & 100.00 & 35.11 & 97.56 & 66.67 & 99.89 \\
4 & 99.56 & 94.00 & 99.11 & 99.67 & 99.56 & 94.00 & 99.11 & 99.67 & 99.56 & 98.44 & 99.67 & 99.56 \\
5 & 100.00 & 97.21 & 100.00 & 100.00 & 66.67 & 94.00 & 99.11 & 99.67 & 66.67 & 98.44 & 99.67 & 99.56 \\
6 & 99.66 & 92.33 & 95.40 & 99.15 & 66.56 & 94.00 & 63.67 & 99.67 & 66.44 & 98.44 & 66.22 & 99.56 \\
7 & 100.00 & 100.00 & 100.00 & 100.00 & 100.00 & 100.00 & 100.00 & 100.00 & 99.89 & 97.22 & 99.67 & 99.78 \\
8 & 100.00 & 99.83 & 100.00 & 100.00 & 66.67 & 100.00 & 66.67 & 100.00 & 66.67 & 97.22 & 66.67 & 99.78 \\
9 & 100.00 & 100.00 & 100.00 & 100.00 & 66.67 & 100.00 & 66.44 & 100.00 & 67.22 & 97.22 & 66.67 & 99.78 \\
10 & 100.00 & 98.78 & 99.22 & 100.00 & 100.00 & 98.78 & 99.22 & 100.00 & 100.00 & 97.44 & 99.56 & 100.00 \\
11 & 100.00 & 100.00 & 100.00 & 100.00 & 66.67 & 98.78 & 66.67 & 100.00 & 66.78 & 97.44 & 66.67 & 100.00 \\
12 & 100.00 & 98.26 & 100.00 & 100.00 & 66.67 & 98.78 & 66.67 & 100.00 & 36.67 & 97.44 & 66.89 & 100.00 \\ \hline
\multicolumn{13}{c}{ISU} \\ \hline
 & \multicolumn{4}{c}{Scenario 1} & \multicolumn{4}{c}{Scenario 2} & \multicolumn{4}{c}{Scenario 3} \\ \cline{2-13} 
 & Local training & FedAvg & FedCos & FedSNGP & Local training & FedAvg & FedCos & FedSNGP & Local training & FedAvg & FedCos & FedSNGP \\
1 & 100.00 & 98.00 & 100.00 & 100.00 & 50.00 & 98.30 & 50.00 & 100.00 & 52.35 & 90.85 & 50.00 & 100.00 \\
2 & 100.00 & 99.50 & 100.00 & 100.00 & 50.35 & 98.30 & 50.00 & 100.00 & 53.60 & 90.85 & 50.00 & 100.00 \\
3 & 100.00 & 99.10 & 100.00 & 100.00 & 50.20 & 98.30 & 50.00 & 100.00 & 55.85 & 90.85 & 50.00 & 100.00 \\
4 & 100.00 & 99.70 & 100.00 & 100.00 & 50.00 & 99.20 & 50.00 & 100.00 & 50.15 & 100.00 & 50.00 & 100.00 \\
5 & 100.00 & 100.00 & 100.00 & 100.00 & 50.00 & 99.20 & 50.00 & 100.00 & 54.40 & 100.00 & 50.00 & 100.00 \\
6 & 100.00 & 98.70 & 100.00 & 100.00 & 50.00 & 99.20 & 50.00 & 100.00 & 50.25 & 100.00 & 50.00 & 100.00 \\
7 & 100.00 & 98.10 & 100.00 & 100.00 & 50.00 & 96.75 & 50.00 & 100.00 & 24.75 & 80.95 & 50.00 & 98.65 \\
8 & 100.00 & 95.80 & 100.00 & 100.00 & 50.05 & 96.75 & 50.00 & 100.00 & 26.60 & 80.95 & 49.85 & 98.65 \\
9 & 100.00 & 99.60 & 99.90 & 100.00 & 50.25 & 96.75 & 49.95 & 100.00 & 27.45 & 80.95 & 49.90 & 98.65 \\
10 & 100.00 & 99.60 & 100.00 & 100.00 & 50.00 & 98.55 & 50.00 & 100.00 & 50.20 & 99.40 & 50.00 & 100.00 \\
11 & 100.00 & 99.70 & 100.00 & 100.00 & 50.40 & 98.55 & 50.00 & 100.00 & 50.25 & 99.40 & 50.00 & 100.00 \\
12 & 100.00 & 97.80 & 100.00 & 100.00 & 50.05 & 98.55 & 50.00 & 100.00 & 53.25 & 99.40 & 50.00 & 100.00 \\ \hline
\end{tabular}%
}
\end{table}
}

\begin{document}

\maketitle
\begin{abstract}
Operators from various industries have been pushing the adoption of wireless sensing nodes for industrial monitoring, and such efforts have produced sizeable condition monitoring datasets that can be used to build diagnosis algorithms capable of warning maintenance engineers of impending failure or identifying current system health conditions. However, single operators may not have sufficiently large fleets of systems or component units to collect sufficient data to develop data-driven algorithms. Collecting a satisfactory quantity of fault patterns for safety-critical systems is particularly difficult due to the rarity of faults. One potential solution to overcome the challenge of having limited or not sufficiently representative datasets is to merge datasets from multiple operators with the same type of assets. This could provide a feasible approach to ensure datasets are large enough and representative enough. However, directly sharing data across the company's borders yields privacy concerns. Federated learning (FL) has emerged as a promising solution to leverage datasets from multiple operators to train a decentralized asset fault diagnosis model while maintaining data confidentiality. However, there are still considerable obstacles to overcome when it comes to optimizing the federation strategy without leaking sensitive data and addressing the issue of client dataset heterogeneity. This is particularly prevalent in fault diagnosis applications due to the high diversity of operating conditions and system configurations. To address these two challenges, we propose a novel clustering-based FL algorithm where clients are clustered for federating based on dataset similarity. To quantify dataset similarity between clients without explicitly sharing data, each client sets aside a local test dataset and evaluates the other clients' model prediction accuracy and uncertainty on this test dataset. Clients are then clustered for FL based on relative prediction accuracy and uncertainty. Experiments on three bearing fault datasets, two publicly available and one newly collected for this work, show that our algorithm significantly outperforms FedAvg and a cosine similarity-based algorithm by $5.1\%$ and $30.7\%$ on average over the three datasets. Further, using a probabilistic classification model has the additional advantage of accurately quantifying its prediction uncertainty, which we show it does exceptionally well. 
\end{abstract}

\blfootnote{Manuscript submitted to \textit{Mechanical Systems and Signal Processing}}

\section{Introduction}
\label{sec:introduction}
The increased availability of sensor data from fleets of cloud-connected assets, such as vehicles and manufacturing facilities, has been driving a transformation in system health monitoring for fault diagnosis. Plant operators and process engineers are interested in leveraging their data to proactively diagnose faulty equipment and notify maintenance personnel of impending potential failures so they can schedule accordingly, improving reliability and reducing downtime costs in the process \citep{cerrada2018review,lee2014prognostics}. To facilitate this effort, numerous data-driven diagnosis algorithms have been developed, enabling real-time detection and accurate classification of system faults \citep{liu2022machine,rao2022review}. Data-driven fault diagnosis methods rely on machine learning and deep learning models to effectively classify condition monitoring signals that indicate the system's health status. Numerous fault diagnosis models have been proposed previously, including artificial neural networks \citep{lin2022gear,shen2021physics, wang2020missing}, random forests \citep{hu2020rotating, ma2018high}, and support vector machines \citep{zhang2015intelligent,wang2020rolling}. In general, the performance of data-driven diagnosis models is directly related to the quality and quantity of available training data \citep{cerrada2018review}. However, the fault patterns that data-driven models are trained to recognize are often unique to the specific system they were collected from and are highly influenced by the operating conditions. This presents a significant challenge to model development since collecting training data under all possible working conditions is time-consuming and costly, which impedes the deployment of data-driven diagnosis models to the field \citep{fink2020potential}. A potential solution to tackle the challenges posed by fault rarity and high variability in operating conditions could be to compile data from multiple clients into a central database for training. Most clients are hesitant to do so because of data privacy concerns and legal regulations \citep{chen2021federated}. In turn, there is a great need to develop decentralized machine learning algorithms that maintain data privacy while providing greater fault classification accuracy than asset operators can achieve individually \citep{fink2021artificial}.

In 2017, McMahan et al. \citep{mcmahan2017communication} developed the first federated learning (FL) strategy for training a centralized deep learning model to serve multiple clients without explicitly requiring them to share their data, thus preserving privacy. In this method, each client trains a model locally with its own data and uploads its model parameters to a server. Then, the server takes a weighted average of each client's model parameters, where the weights are determined by the size of the client's training dataset. This method of model aggregation on the server has since been referred to as federated averaging (FedAvg). Several researchers have effectively applied FedAvg and similar methods to train fault diagnosis models \citep{zhang2021adaptive, xia2022privacy, zhang2021federated, zhang2023blockchain}. Notably, Zhang et al. \citep{zhang2021federated} demonstrated a dynamic validation and self-supervised FedAvg algorithm where the server evaluated local models using a validation dataset to determine which models to neglect during federation. A follow-up work by Zhang et al. \citep{zhang2023blockchain} introduced blockchain technology to improve the security of model parameter sharing. While FedAvg has shown promise in cross-client fault diagnosis, challenges arise when clients' data are heterogeneous or non-independent and identically distributed (non-IID) \citep{niknam2020federated, li2019convergence, zhu2021federated}. 

The performance of FL algorithms can be affected by two types of statistical data heterogeneity. The first, known as \emph{domain shift}, occurs when clients' data do not share a similar distribution due to differences in system operating conditions or system configurations. The second, referred to as \emph{label heterogeneity}, occurs when clients' datasets have different quantities and types of system faults \citep{zhu2022aligning}. In most real-world scenarios, clients' data generally do not exhibit similar distributions, fault types, and fault prevalence due to the diverse environmental conditions experienced by systems or components in the field (e.g., run-time, load, temperature, and humidity) \citep{fink2021artificial}. Unfortunately, large variations between clients' datasets have been shown to reduce the training speed and overall accuracy of the FedAvg algorithm \citep{li2019convergence, wu2022node}. One potential solution to deal with domain shift among client's datasets is to develop a personalized feature extractor network that extracts fault-related features that are invariant to the different operating conditions and fault types of the clients (see Fig. \ref{fig:Intro_FL}b) \citep{kevin2021federated,luopan2022fedknow,wang2022novel}. This approach, referred to as adaptation-based transfer learning, mitigates the data domain shift using unique model architectures that extract features that are indistinguishable between different clients. Another approach to deal with domain shift is to group clients for training based on their dataset similarity, where each group builds a different global model (see Fig. \ref{fig:Intro_FL}c) \citep{li2022clustering,li2021federated,wang2021adaptive,tian2022wscc}. The key to this approach lies in the clustering strategy used to group the clients since directly comparing clients' datasets would violate data privacy. For example, Tian et al. \citep{tian2022wscc} utilized the cosine similarity between each pair of client model parameters to infer the similarity of client dataset distributions. While both transfer learning and clustering are effective approaches to deal with client data heterogeneity, numerous distinct challenges specific to system fault diagnosis must be resolved.

Firstly, due to the differences in operating settings and monitoring policies, the frequency of fault types presented in clients' datasets can be significantly heterogeneous, and some fault types may be completely missing. Few studies have addressed the challenge of training a model algorithm that can accurately differentiate all labels in scenarios where specific labels are missed in certain clients. Second, most studies adopt a deterministic model for machinery health monitoring. The deterministic models used in those studies are capable of addressing data heterogeneity challenges under the assumption that the test data follows a similar distribution as the training data. However, in a real-world application, models may be faced with out-of-distribution samples. Considering that scenario, it is desirable to use a probabilistic model that can accurately represent its prediction uncertainty and confidence in the prediction results. Considering the aforementioned requirements, building a cross-client fault diagnosis model that can be deployed in practice is undoubtedly difficult. To meet the needs of the industry, it is important to use a FL strategy that can effectively train a fault diagnosis model even when many clients are missing one or more fault types. Furthermore, the final global model distributed to each client should output accurate uncertainty estimates with its fault diagnosis predictions so that maintenance personnel can appropriately gauge the severity of the situation.

In this work, we develop a dynamic clustering FL algorithm for training an accurate fault diagnosis model even when clients' datasets exhibit significant label heterogeneity and domain shifts. Different from existing algorithms that use model parameters or prediction errors to cluster clients, we use a probabilistic deep learning model and leverage its prediction uncertainty estimates to cluster clients based on inferred dataset similarity. We demonstrate the ability of our uncertainty-based clustering strategy to assign clients with similar data distributions to the same cluster for training. We benchmark the fault classification performance of our federated learning algorithm (FedSNGP) against FedAvg and a cosine similarity-based clustering algorithm (FedCos) using two publicly-available bearing fault datasets and one new bearing fault dataset collected specifically for this work. In cases where clients' datasets are heterogeneous in fault type and exhibit significant domain shift (Scenario 2), our FedSNGP algorithm outperforms FedAvg and FedCos. Further, in cases where clients' datasets exhibit domain shift and are heterogeneous in fault type as well as the sample size (Scenario 3), our FedSNGP algorithm demonstrates its prowess, achieving significantly higher accuracy than both FedAvg and FedCos. This final scenario highlights the advantage of our self-supervised and uncertainty-aware clustering algorithm for FL, as it achieves excellent fault classification accuracy, even in the most extreme scenarios. Further, using a probabilistic classification model has the added benefit of quantifying its prediction uncertainty to detect out-of-distribution samples, which we show it does exceptionally well.

\section{Preliminaries on FL}
\label{sec:preliminaries_on_federated_learning}

\subsection{Basics of FL}

This study targets industrial scenarios where each operator has its own dataset and shares the same diagnosis task. To stay consistent with existing literature on federated learning, we refer to operators as clients. FL is a distributed machine learning approach for training a model using decentralized data from multiple clients without sharing or aggregating the clients' datasets on a central server. We begin by defining notations used to describe FL strategies. We denote the number of clients as $N$  and denote the $i^\mathrm{th}$ client’s local training dataset as: $D^{i} = \left\{ \left( {\mathbf{x}_{j}^{i},y_{j}^{i}} \right) \right\}_{j = 1}^{n_{i}}$, where each client has $n_{i}$ input and output training data pairs, denoted as $\mathbf{x}_{j}^{i}$ and $y_{j}^{i}$, respectively. In the traditional centralized model training approach, the first involves aggregating all the client's data on a single server, followed by optimizing each client's model parameters $\boldsymbol{\uptheta}^{i}$ to minimize the global loss $\sum\limits_{i = 1}^{N}{\sum\limits_{j = 1}^{n_{i}}{Loss\left( {\boldsymbol{\uptheta}^{i},\mathbf{x}_{j}^{i},y_{j}^{i}} \right)}}$. Here,
$Loss\left({\boldsymbol{\uptheta}^{i},\mathbf{x}^{i}_{j},y^{i}_{j}} \right)$ represents the prediction loss of the model for a given input-output pair $\left(\mathbf{x}^{i}_{j},y^{i}_{j} \right)$ with respect to the model parameters $\boldsymbol{\uptheta}^{i}$. 

\begin{figure}[h] 
	\includegraphics[width=1\textwidth]{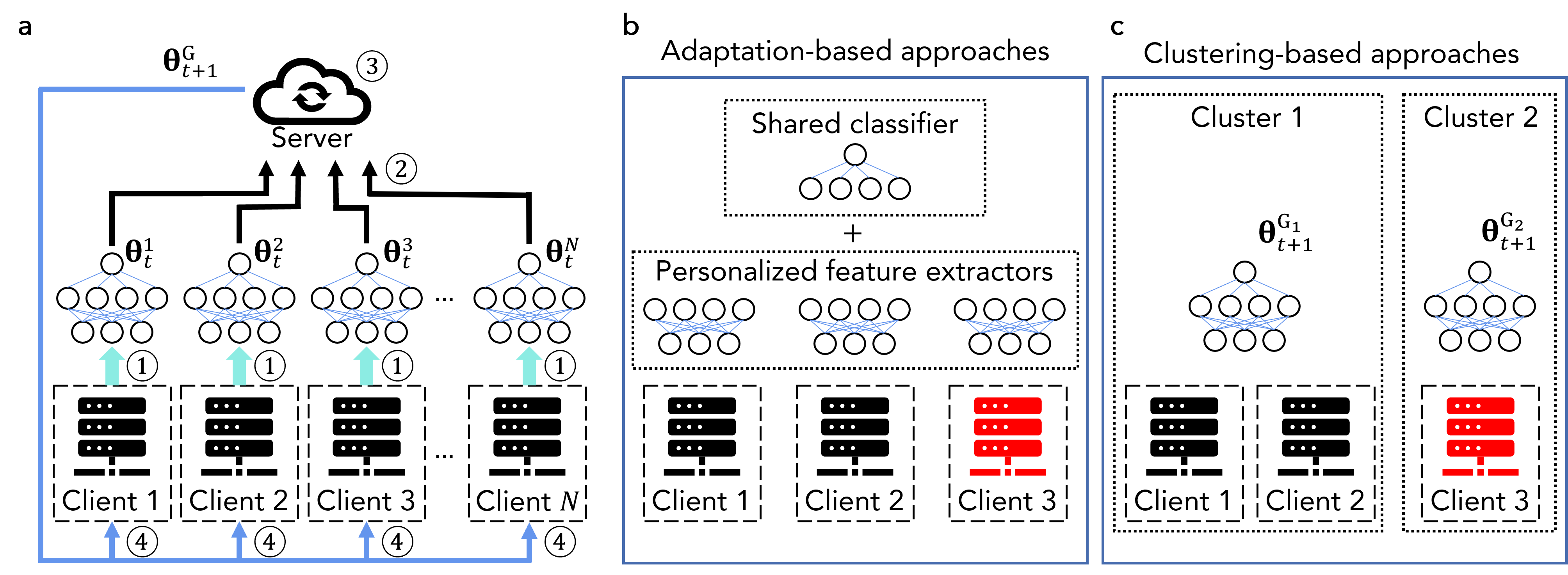}
	\centering
	\caption{An illustration of FL (\textbf{a}) and two FL strategies (\textbf{b}, \textbf{c}). \textbf{a}, The four-step process of FL is common to nearly all FL approaches. \textbf{b}, An overview of an adaptation-based approach to FL that uses a global classifier to accommodate client-specific feature extractors. \textbf{c}, An overview of a clustering-based approach to FL that groups clients for training based on non-private comparison criteria.}
	\label{fig:Intro_FL}
	\centering
\end{figure}

One simplistic approach is to allow the server to aggregate all of the client datasets and optimize the model parameters by computing the gradient of the global loss. This implies that the server has complete access to all of the clients' datasets. However, aggregating sensitive client data on a central server presents privacy and legal concerns and should be avoided. Instead, FL strategies aim to train a global model or models without sharing data between clients. To perform FL, as illustrated in Fig. \ref{fig:Intro_FL}a, each client independently optimizes their local model using their own dataset (Step 1) and transmits their model parameters to the server (Step 2). On the server side, model parameters are aggregated to create one or more global models (Step 3), which are subsequently disseminated  to the clients (Step 4). A single instance of this four-step process is often called a \emph{communication round}, as it involves the clients communicating with the server to facilitate training. 

The most important step in any FL strategy is the model aggregation (Step 3 in Fig. \ref{fig:Intro_FL}a). Model aggregation leverages non-sensitive client information (e.g., model parameters and other information the client is willing to share) as part of an algorithm to update the global model. The model aggregation algorithm directly affects the amount of information transferred from each client to the final model and is crucial to the model's overall accuracy. The first model aggregation strategy, proposed by McMahan et al. \citep{mcmahan2017communication}, was the FedAvg strategy. The FedAvg method of model aggregation combines the client's models by performing a weighted average of model parameters, where the weights are determined based on the number of training samples each client has. The weighted average model aggregation is defined as follows:
\begin{equation}\label{Eqn:Weighted_average}
    \boldsymbol{\uptheta}_{t + 1} = \frac{\sum_{i = 1}^{N}{n_{i}\boldsymbol{\uptheta}_{t}^{i}}}{\sum_{i = 1}^{N}n_{i}}
\end{equation}
The weighted averaging approach to model aggregation used in FedAvg allows clients to employ a collaborative prediction model that contains additional information beyond what each client had separately. This is because each client's model can be regarded as a condensed representation of their data, and constructing a global model by incorporating the weights of all clients enables the global model to learn each client’s information indirectly. As a result, this yields an exceptionally accurate model for the clients. 

\subsection{FL for Non-IID Datasets}

Though FedAvg yields decent performance in most scenarios, a major challenge of the FedAvg approach is that its performance decreases significantly when the clients' data are vastly different from one another, i.e., non-IID and heterogeneous in fault types \citep{fink2020potential,li2019convergence}. Large differences in local dataset distributions (often referred to as domain shifts) cause some clients' model parameters to deviate significantly from those of the group. Including information models from clients with significantly different dataset distributions can negatively impact the global model aggregation process, causing the final global model to serve neither the majority of clients with similar data nor the minority clients with outlier data well. Numerous algorithms have been developed to tackle the challenge of data heterogeneity when deploying a FL strategy for a fleet of system or component units. This section introduces two of the most common methods for dealing with non-IID datasets and highlights notable research in this area.

\subsubsection{Adaptation-based Approaches to FL}

The challenge of data domain shift has been widely recognized in the field of deep learning  \citep{li2022perspective}. One common method of dealing with data domain shift in traditional deep learning settings is known as adaptation-based learning \citep{tan2022towards,zhu2022aligning}. There are different domain adaptation approaches. One way is to train a feature extractor that extracts domain-invariant features from the input data. Then, feed the domain invariant features to a classifier for prediction. 

The same domain adaptation approach can be applied to the FL problem. In adaptation-based FL, shown in Fig. \ref{fig:Intro_FL}b, the $i^\mathrm{th}$ client's model is divided into a personalized feature extractor $\varphi_{i}$ and a shared global classifier $h_{i}$\citep{wang2022novel,zhu2022aligning}. The personalized feature extractor learns to map the raw input data into a low-dimensional feature space such that the learned mapping is invariant to the system's operating conditions. Then, the shared classifier takes the extracted domain-invariant features as input to diagnose the system's health. Mathematically, this optimization process can be written as follows:

\begin{equation}\label{Eqn:Adaptation_FL}
\begin{matrix}
{\textrm{argmin}} \\
{\varphi_{1},\varphi_{2},\ldots,\varphi_{N,h_{g}}} \\
\end{matrix}{\sum\limits_{i = 1}^{N}{\sum\limits_{j = 1}^{n_{i}}{Loss\left( {\varphi_{i} \circ h_{g},\mathbf{X}_{j}^{i},y_{j}^{i}} \right)}}} + \alpha{\sum\limits_{i = 1}^{N}{\sum\limits_{j = 1,j \neq i}^{N}{Dis\left( {z_{i},z_{j}} \right)}}}
\end{equation}
where ${Loss\left( {\varphi_{i} \circ h_{g},\mathbf{X}_{j}^{i},y_{j}^{i}} \right)}$ denotes the $i^\mathrm{th}$ model's prediction loss towards the local training dataset, $z_{i}$ denotes the client $i$'s extracted features, and $Dis(\cdot)$ is the metric of discrepancy of two distributions.

Training a personalized feature extractor can be achieved by introducing a domain classifier into the model architecture \citep{wang2022novel} or introducing the maximum mean discrepancy distance into the loss function \citep{chen2022federated}. In the context of an FL framework where new clients may join, researchers in \citep{chen2022federated} developed a dynamic weighted averaging algorithm. In the model aggregation step, the algorithm evaluates the maximum mean discrepancy between features extracted from source domain models and the target domain model. A smaller discrepancy implies that there is less domain discrepancy between two clients and should contribute more to the training of the target domain model. The algorithm alleviates domain discrepancy by assigning higher weights to models that can extract similar features in the source and target domains.

While domain adaptation algorithms are a popular solution for dealing with data domain shifts, their main limitation is their lack of flexibility to changes in future operating conditions. Domain adaptation algorithms build personalized feature extractors for each client (target domain) and assume that the data distribution and operating conditions will not change with time. If any aspect of the system changes, the input data will shift, and the personalized feature extractor cannot be trusted to accurately map the new inputs into domain-invariant features. To alleviate this challenge, test-time adaptation algorithms have been developed. Regardless, domain adaptation algorithms remain promising to deal with domain shifts in FL environments.

\subsubsection{Clustering-based Approaches to FL}

Research conducted in \citep{chen2021federated, zhang2021federated} has highlighted that in non-IID FL scenarios, certain clients should be disregarded or assigned lower weights during model aggregation to prevent the transfer of low-quality knowledge to the global model. During each training communication round of clustering-based FL, the server performs client clustering that assigns clients with comparable data distributions to the same cluster, whereby the clients in the same cluster share the same model parameters. The optimization target of the clustering strategy and model parameter optimization can be formulated as:
\begin{equation}\label{Eqn:Clustering_FL}
 \begin{aligned}
\begin{matrix}
{\textrm{minimize}} \\
{C_{1},C_{2},\ldots C_{k}} \\
\end{matrix}\frac{1}{N}{\sum\limits_{k = 1}^{K}{\sum\limits_{i = 1}^{N}{r_{i,k}{\sum\limits_{j = 1}^{n_{i}}{Loss\left( {C_{k},\mathbf{X}_{j}^{i},y_{j}^{i}} \right)}}}}}\\
\textrm{subject to } r_{i,k} = \begin{matrix}
{\textrm{argmin}} \\
r_{i,k} \\
\end{matrix}\frac{1}{N}{\sum\limits_{k = 1}^{K}{\sum\limits_{i = 1}^{N}{r_{i,k}\left\| {\mathbf{D}^{i} - \Omega_{k}} \right\|}}}
\end{aligned}
\end{equation}
Where $C_{k}$ denotes the model parameters of the clients in $k^\mathrm{th}$ cluster, $r_{i,k}$ denotes the resulting client clustering with $r_{i,k} =1$ if client $i \in$ cluster $k$, otherwise $r_{i,k} =0$. The $\mathbf{D}^{i}$ denotes client dataset distribution, and $\Omega_{k}$ represents the data distribution center for cluster $k$. It is important to note that in order to maintain data privacy, the server must indirectly deduce dataset similarity ( $\left\| {\mathbf{D}_{i} - \Omega_{k}} \right\|$).

The key to clustering-based FL algorithms is to estimate the dataset similarity between each pair of clients without accessing the client's dataset. Researchers in \citep{tian2022wscc} developed a similarity-based clustering approach that used the cosine distance between models' parameters to estimate their dataset similarity and determine if they should be clustered together for training. In this approach, clients with identical or highly similar dataset distributions are clustered together for each FL communication round. In a similar line of research,  \citet{li2021federated} proposed a soft clustering algorithm that can assign clients to overlapped clusters. In this way, each client's information is fully utilized. 

There is currently limited research that focuses on employing clustering-based algorithms for FL in the context of system fault diagnosis. Researchers in  \citep{gholizadeh2022federated} developed a clustering-based algorithm where each client performs a GridSearchCV method to tune their model hyperparameters. Then during model aggregation, the clients that share similar hyperparameters are clustered into the same group to get updated models. Researchers in \citep{li2022clustering} proposed a multi-center clustering-based FL algorithm. They used a k-means-based algorithm to cluster clients. However, the drawback of this approach is that it requires specifying the number of clusters in advance, which can significantly affect performance.

For clustering-based FL algorithms, the clustering strategy plays a crucial role in determining the algorithm's performance. If a client is allocated to an unsuitable cluster, the resulting global model may perform worse than the more basic algorithms, like FedAvg, that incorporate all models in the federation process regardless of data domain shift and label heterogeneity. Altogether, domain-adaptation and clustering-based approaches have been shown to be effective solutions to the challenge of domain shift and label heterogeneity in FL. However, little work has been done to apply these methods in the field of system fault diagnosis.

\section{Proposed FL Method}
\label{sec:Proposed FL Method}
Our approach to FL differs from existing clustering-based FL approaches in two important ways: Firstly, we incorporate a distance-aware model into the FL framework, which enables us to use the prediction uncertainty to infer the dataset similarity and group clients into clusters. The distance-aware model we employ in the proposed FL approach has a unique capability of detecting out-of-distribution input samples, thus preventing the model from making overconfident predictions. Secondly, we use a clustering algorithm that can determine the number of clusters without requiring user-defined parameters. The number of clusters is dynamically adjusted by the clustering algorithm during each communication round based on the dataset similarity inferred by the distance-aware model. This ensures that the client grouping is optimized to achieve the maximum information transfer to the global model without the need for predefining the number of clusters. Our algorithm is designed to work well under a wide range of dataset conditions with various levels of domain shift and missing fault types. 

\begin{figure}[h]
	\includegraphics[width=1\textwidth]{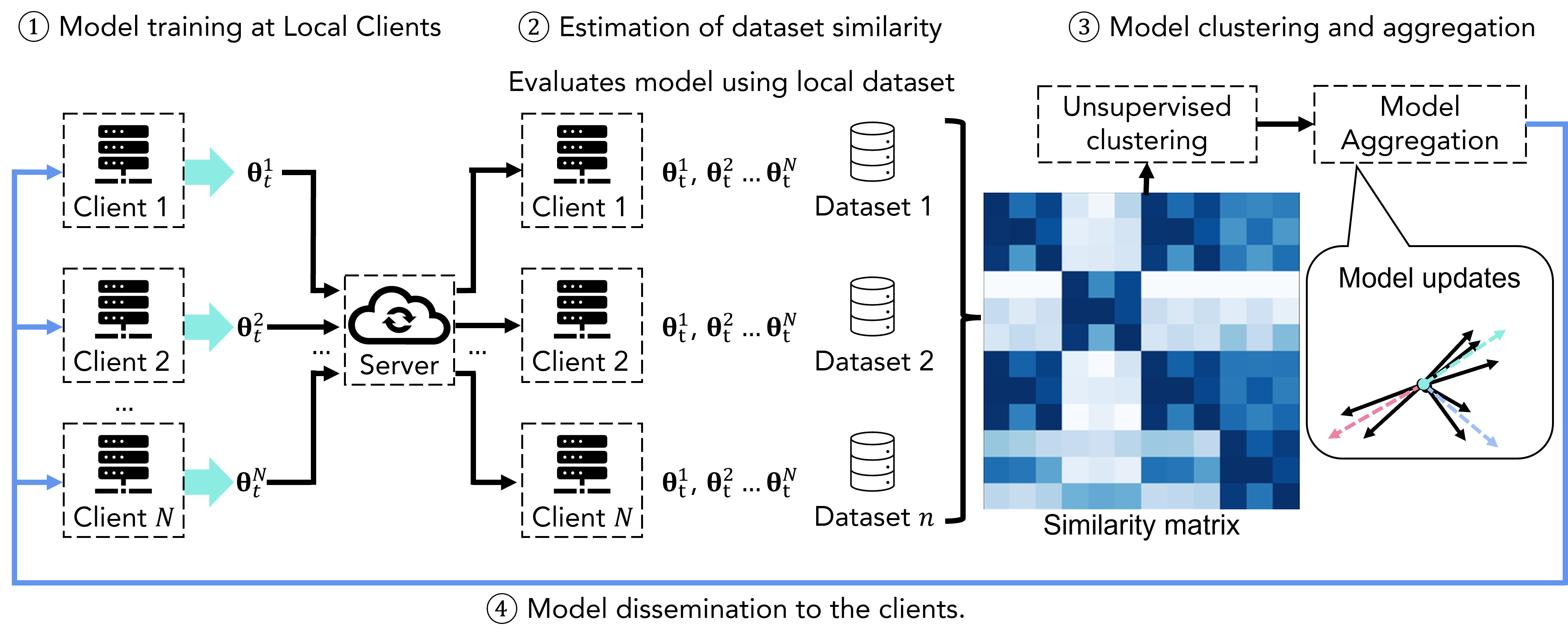}
	\centering
	\caption{An illustration of the FedSNGP training process. Each training communication round has four key steps: 1) local model training, 2) dataset similarity estimation, 3) local model clustering for training, and 4) model dissemination to the clients.} 
	\label{fig:Fedsngp}
	\centering
\end{figure}

Shown in Fig. \ref{fig:Fedsngp}, our FL training strategy consists of four steps: 
\begin{enumerate}
    \item  Local model training: each client trains its own model using its dataset and uploads model parameters to the server. 
    \item Estimation of dataset similarity: each client downloads the models of other clients from the server and estimates the prediction uncertainty of downloaded models towards its dataset to infer distribution similarity between each pair of client datasets. 
    \item Model clustering and aggregation: the server forms a similarity matrix based on the estimated dataset similarity, uses an unsupervised clustering algorithm to group the clients into clusters, and performs model aggregation within each cluster.
    \item Model dissemination to the clients: the server sends the updated global model back to the clients for further local training. 
\end{enumerate}

In what follows, we describe our uncertainty-aware deep learning model and our dynamic clustering algorithm for FL.

\subsection{Uncertainty-Aware Model Training at Local Clients}

Deploying a deep learning model in the field requires building confidence among operators and maintenance personnel regarding the model's diagnostic accuracy. At a minimum, the model should indicate higher predictive uncertainty to operators when it lacks confidence in a prediction. This is particularly important for safety-critical equipment, as an incorrect diagnosis can result in catastrophic failures, putting people and equipment at risk. For this reason, it is essential to build a fault diagnosis model that can accurately estimate its prediction uncertainty. Moreover, we believe that the uncertainty estimates can be leveraged to improve the performance of our FL approach. 

Real-world fault diagnosis is prone to challenges such as domain shifts caused by changes in environmental or operating conditions or other factors. With a slight shift in input distribution, a vanilla machine learning model may yield overconfident prediction results for these samples, misleading operators and wasting time. To tackle this problem, it is desirable to develop an algorithm that can infer when new data exhibit domain shift or unknown fault type with respect to the training dataset. The larger the discrepancy between the new sample and the model's training data, the higher the model's prediction uncertainty should be. To accomplish this, we employ a distinctive model architecture that enables the model to identify input samples that differ significantly from the data it was trained on and consequently increase its uncertainty estimate.

The ability of the model to detect the difference between new samples and the training dataset is referred to as ‘distance awareness’ \citep{liu2020simple}. Given a deep learning model $logit\left( \mathbf{x} \right) = g \circ h\left( \mathbf{x} \right)$, where $g$ represents the model output layer and $h( \cdot )$ denotes the mapping function that maps the raw input into a hidden representation space, a deep learning model is input distance aware if it follows two conditions:

\begin{enumerate}
  \item The model's mapping function is distance-preserving. Considering the distance in the data manifold $\left\| {\mathbf{x} - \mathbf{x}^{'}} \right\|_{X}$, the distance in the hidden space $\left\| {h\left( \mathbf{x} \right) - h\left( \mathbf{x}^{'} \right)} \right\|_{H}$ needs to satisfy the following bi-Lipschitz condition:
  
\begin{equation}
L_{1}\times\left\| {\mathbf{x} - \mathbf{x}^{'}} \right\|_{X} \leq \left\| {h\left( \mathbf{x} \right) - h\left( \mathbf{x}^{'} \right)} \right\|_{H} \leq L_{2}\times\left\| {\mathbf{x} - \mathbf{x}^{'}} \right\|_{X}~,~0 < L_{1} < L_{2}
\end{equation}
  
  \item The model's output layer is distance aware. The output layer $g$ should output an uncertainty metric that reflects distance in the hidden space $\left\| {h\left( \mathbf{x} \right) - h\left( \mathbf{x}^{'} \right)} \right\|_{H}$. The standard Gaussian process machine learning model exhibits this property. However, Gaussian processes are unsuitable for high-dimensional problems and large datasets because the inference is extremely computationally expensive $(\mathcal{O}(n^3))$ \citep{gardner2018gpytorch}.
\end{enumerate}

To address the above-mentioned conditions, Liu et. al \citep{liu2020simple} proposed a model architecture known as the Spectral-normalized Neural Gaussian Process (SNGP). The SNGP is formed by residual blocks equipped with spectral normalization to ensure that the hidden mapping $h( \cdot )$ is distance preserving. For the model's output layer,  the dense layer is replaced with an approximate Gaussian process conditioned on the learned hidden representations of the model. The SNGP adopts random feature expansion to convert the infinite-dimensional Gaussian process into a featured Bayesian linear model and uses Laplace approximation to approximate the posterior. The output layer is designed with fixed hidden weights $\mathbf{w}_{L}$ and trainable output weights $\boldsymbol{\beta}$, given as:
\begin{equation}
g\left( \mathbf{h} \right) = \sqrt{\frac{2}{D_{L}}}cos\left( {- \mathbf{w}_{L}\mathbf{h} + \mathbf{b}_{L}} \right)^{\intercal}\boldsymbol{\beta}, \mathrm{with~prior~} \boldsymbol{\beta}_{D_{L}\times1} \sim N\left( 0,\mathbf{I}_{D_{L} \times D_{L}} \right)
\end{equation}
where $\mathbf{h}$ denotes the output features of the penultimate layer with dimension $D_{L - 1}$, $\mathbf{w}_{L}$ is a fixed weight matrix with dimension $D_{L} \times D_{L - 1}$ whose entries are sampled i.i.d. from $\sim N\left( 0,1 \right)$. $\mathbf{b}_{L}$ is a fixed bias term with dimension $D_{L} \times 1$ whose entries are sampled i.i.d. from $Uniform(0,\pi)$. Conditional on $\mathbf{h}$, $\boldsymbol{\beta}$ is the only trainable variable in the output layer. For K-class classification, the Laplace method is applied to approximate the posterior for each class as \citep{liu2020simple}:
\begin{equation}
p(\boldsymbol{\beta}_{k}\mid D) \approx MVN\left( {{\hat{\boldsymbol{\beta}_{k}}},{\hat{\Sigma}}_{k} = {\hat{\mathbf{H}_{k}}}^{- 1}} \right)~, \mathrm{where} ~{\hat{\mathbf{H}}}_{k} = {\sum\limits_{i = 1}^{N}{p_{i,k}\left( {1 - p_{i,k}} \right)\Phi_{i}\Phi_{i}^{\intercal} }}+\mathbf{I}
\end{equation}
and ${\hat{\boldsymbol{\beta}}_{k}}$  is the model's maximum a posteriori probability estimate conditioned on the random Fourier hidden representation, $p_{i,k}= 1/(1 + e^{- \Phi_{i}^{\intercal}\hat{\boldsymbol{\beta}_{k}}})$, and $\Phi_{i} = \sqrt{\frac{2}{D_{L}}}*cos\left( - \mathbf{w}_{L}h(\mathbf{x}_{i}) + \mathbf{b}_{L} \right)$.

Training the SNGP model can be performed using stochastic gradient descent:
\begin{equation}
-\log p\left(\boldsymbol{\beta},\left\{\mathbf{w}_{l}, \mathbf{b}_{l}\right\}_{l=1}^{L-1} \mid \mathcal{D}\right)=-\log p\left(\mathcal{D} \mid \boldsymbol{\beta},\left\{\mathbf{w}_{l}, \mathbf{b}_{l}\right\}_{l=1}^{L-1}\right)+\frac{1}{2}\|\boldsymbol{\beta}\|^{2}
\end{equation}
where $-\log p\left(\mathcal{D} \mid \boldsymbol{\beta},\left\{\mathbf{w}_{l}, \mathbf{b}_{l}\right\}_{l=1}^{L-1}\right)$ is the cross-entropy loss for the classification task. The pseudo-code of the training and test procedure of the SNGP model is summarized in \textbf{Algorithms 1 and 2}. Each client trains a local model using the local dataset and uploads the model to the server.

\begin{figure}[h] 
	\includegraphics[width=0.7\textwidth]{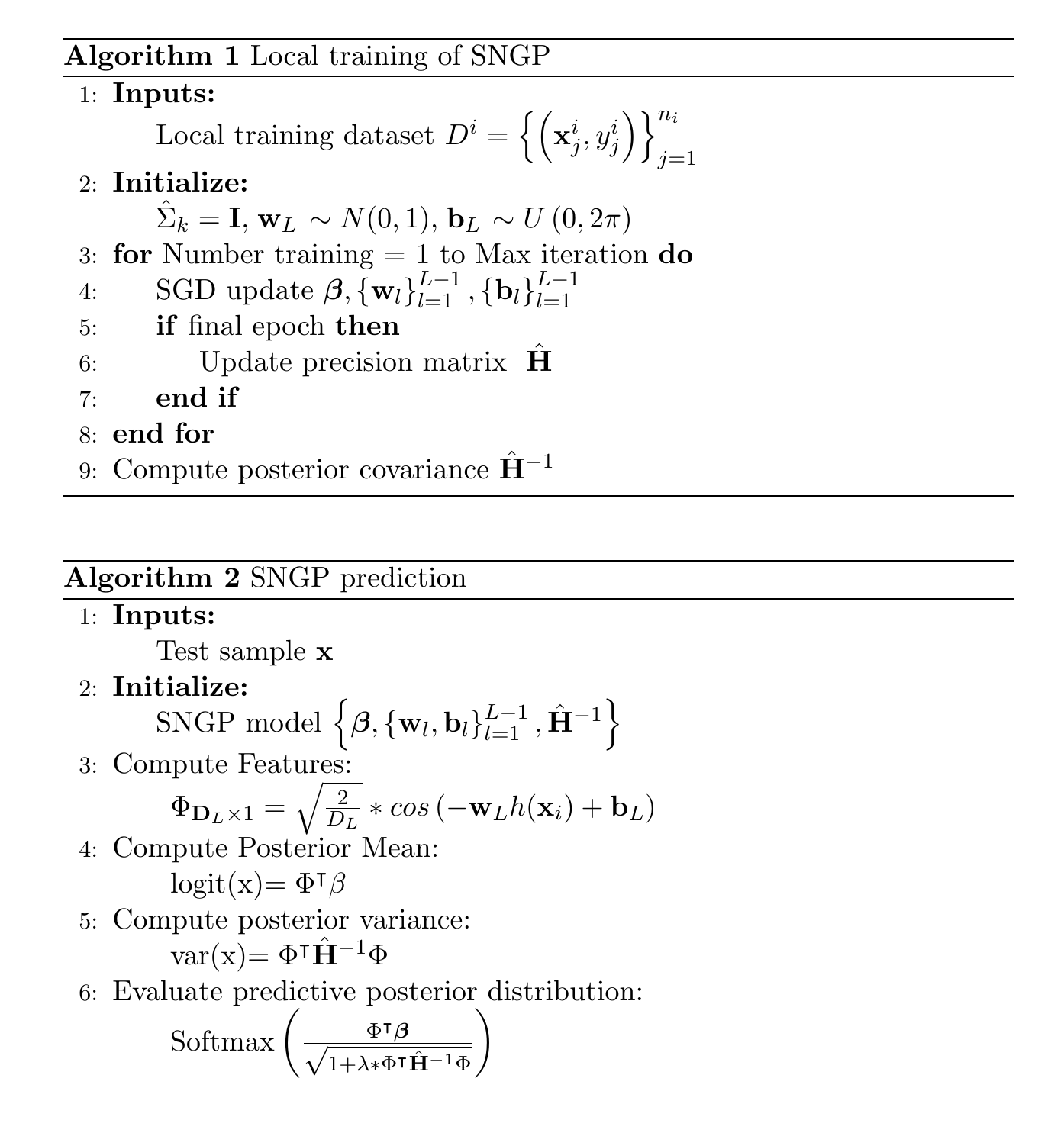}
	\centering
  	\caption{Pseudo-code of the training and prediction process of the SNGP model}
	\label{table:Algorithm1and2}
	\centering
\end{figure}

\subsection{Dataset Similarity Estimation for Federated Clustering}

When dealing with client datasets that are heterogeneous, the typical methods of FL strategies may result in suboptimal performance. One way to alleviate the negative impact caused by data heterogeneity is to cluster clients \citep{han2022survey}. Clustering-based FL algorithms group clients into clusters based on their dataset similarity. The SNGP model is capable of inferring the distance between the training dataset and test samples. Therefore, SNGP model can be adopted to infer the similarity between the datasets of two clients. The FedSNGP algorithm involves uploading the locally trained model from each client to the server after training. Following this, the clients work together to estimate the similarities between their datasets.

As the SNGP model is distance aware, the model's prediction result reflects the similarity between the test sample and the training datasets. To infer client dataset similarity, each client downloads other clients' models and evaluates other models' prediction results against the local training dataset. The SNGP model $\left\{ \boldsymbol{\beta},\left\{\mathbf{w}_{l}, \mathbf{b}_{l}\right\}_{l=1}^{L-1},\hat{\mathbf{H}}^{-1}\right\}$ maps the input $\mathbf{x}$ into random Fourier feature $
\Phi = \sqrt{\frac{2}{D_{L}}}cos\left( \mathbf{w}_{L}h\left( \mathbf{x} \right) + \mathbf{b}_{L} \right)$ , and the output posterior predictive probability for each class follows $MVN\left( {\Phi^{\intercal}\boldsymbol{\beta}_{k},\Phi^{\intercal}\hat{\mathbf{H}}^{-1}_{k}\Phi} \right)$, the expectation of the probability can be approximated by using the mean-field method:
\begin{equation}
E\left( p\left( \mathbf{x} \right) \right) \approx \text{Softmax}\left( \frac{\Phi^{\intercal}\boldsymbol{\beta}_{k}}{\sqrt{1 + \lambda*\Phi^{\intercal}{\hat{\mathbf{H}_{k}}^{-1}}\Phi}} \right)
\end{equation}
where $\lambda$ is a scaling factor that is commonly set to $\pi/8$ \citep{liu2020simple,lu2020mean}. Then, in this study, the predicted variance is used to infer model uncertainty: 
\begin{equation}
P_{var} \approx \frac{1}{n_{i}~}{\sum\limits_{i = 1}^{n_{i}}{var\left( {p\left( \mathbf{x}_{\mathbf{i}} \right)} \right)}} = \frac{1}{n_{i}~}{\sum\limits_{i = 1}^{n_{i}}{E\left( {p\left( \mathbf{x}_{\mathbf{i}}^{2} \right)} \right) - \left( {E\left( {p\left( \mathbf{x}_{i} \right)} \right)} \right)^{2}}}
\end{equation}
The predicted variance reflects the distance between a model’s training and test datasets, where higher variance indicates a higher difference between two datasets. In the proposed FedSNGP approach, each client downloads all the local models and evaluates those models using its local training dataset. Then, the predicted variance results are uploaded to the server and are used for clustering. The predicted variance is used to form a 2D matrix, where the entry in $i^\mathrm{th}$ row, $j^\mathrm{th}$ column is the predicted variance from evaluating the $j^\mathrm{th}$ model using the $i^\mathrm{th}$ client's dataset. After performing column-wise normalization, the results form a prediction uncertainty matrix with the range $[0,1]$, with dimension $N_{client} \times N_{client}$, denoted as $M_{U}$.

In the next step, the affinity propagation algorithm \citep{frey2007clustering} is used to automatically determine the number of clusters. Unlike the k-means clustering algorithm, affinity propagation does not require the user to specify the number of clusters in advance. Instead, it automatically determines a suitable number of clusters. We create a similarity matrix $M_{Sim}$ by calculating $1-M_{U}$. The similarity matrix represents the similarity between clients' datasets, where values close to 0 indicate high prediction uncertainty and suggest that the $j^\mathrm{th}$ client's dataset differs significantly from the $i^\mathrm{th}$ client's dataset, and the clients would not benefit from federating together. The affinity propagation takes the similarity matrix as input and outputs the clustering strategy for model aggregation. The clustering strategy takes place remotely on the central server. The algorithm describing this process is summarized in \textbf{Algorithm 3}.



\begin{figure}[h] 
	\includegraphics[width=0.7\textwidth]{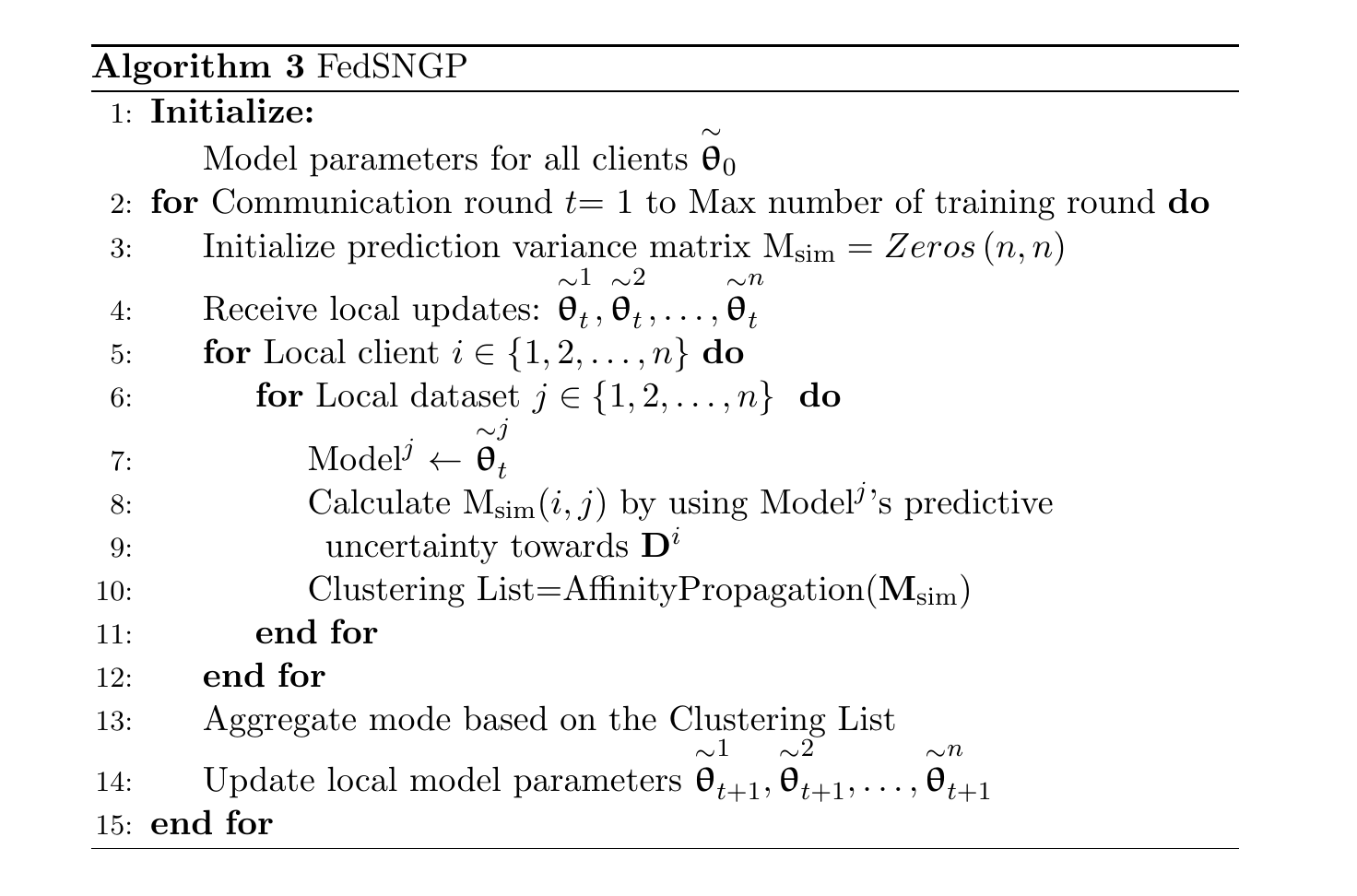}
	\centering
 	\caption{Pseudo-code of FedSNGP algorithm}
	\label{Figure:Algorithm3}
	\centering
\end{figure}

\section{Training and Test Configuration}
\label{sec:Training and Test Configuration}
\subsection{Selected Datasets}

We evaluate the performance of the proposed algorithm on three bearing fault diagnosis case studies to demonstrate the effectiveness of the FedSNGP algorithm. A summary of three case studies is given in Table \ref{table:Dataset}. Each bearing dataset contains vibration signals collected from rolling element bearings with distinct health conditions under various operating conditions. To simulate the FL scenario, the dataset is partitioned into 12 subsets, and each subset is allocated to a client. The objective of each client is to train a fault diagnosis model capable of accurately identifying the health condition of bearings. The selected bearing datasets chosen for this task are as follows:

\begin{enumerate}[leftmargin=*]
\item Case Western Reserve University (CWRU) dataset \citep{CWRU_dataset}: Case study 1 uses the Case Western Reserve University (CWRU) bearing dataset, which is widely used as a standard reference in the bearing diagnosis field. The data was collected with a 12 kHz sampling frequency under four working conditions. The bearing faults were introduced by using electro-discharge machining. Three manual fault sizes were created, with diameters of 0.007, 0.014, and 0.021 inches. Considering that outer race and inner race faults are more frequently appeared in accelerated degradation tests, we use the healthy, inner race fault, and outer race fault bearings. For each type of fault, the bearings with fault diameters of 0.007 and  0.014 inches are selected.

\item Paderborn University (PU) dataset \citep{lessmeier2016condition}: Case study 2 uses an experimental dataset collected by the KAT data center in Paderborn University (PU). The PU dataset contains high-resolution vibration data, which are collected from six healthy bearings and 26 damaged bearings under four working conditions. The sampling frequency is 64kHz. Out of the 26 damaged bearing sets, 12 bearings were artificially damaged, and the other 14 were damaged by accelerated lifetime test. Bearings that undergo accelerated lifetime testing are selected in this case study. In total, bearings with three different health conditions (healthy, inner race fault, and outer race fault) are included, with three bearings for each health condition.

\item  Iowa State University (ISU) dataset:  An experimental dataset is collected from a machinery fault simulator in our lab at Iowa State University, denoted as the ISU dataset. The ISU dataset was collected with a 25.6 kHz sampling frequency from four bearings under four working conditions. The bearing faults were introduced by electrical discharge machining, where the size of each fault is approximately 1.5 mm× 1.0 mm × 0.1 mm. Four health conditions are considered: healthy, inner race fault, outer race fault, and a combination of faults.
\end{enumerate}

\TableOne

The above-mentioned datasets were collected under different sampling rates and different sampling times. We first resample vibration signals with sampling frequency = 12.8 kHz. Then split each signal into multiple vibration samples. The length of each sample is equal to 1024 points and is equivalent to 0.08 seconds of data. Frequency domain analysis has been widely applied in vibration-based bearing diagnosis \citep{shen2021physics,michau2021unsupervised}. In this study, we apply the fast Fourier transform to acquire the single-sided output of the signal power spectrum. After signal processing, each time domain signal is converted into a  frequency domain feature, with length = 512. After dataset processing, 80 \% of the total dataset is used to form training datasets, while the remainder is used for testing.

\subsection{Design of the FL Scenarios}

To demonstrate the impact of dataset heterogeneity,  we considered three FL scenarios in which each client's dataset is collected under different working conditions. Moreover, due to the rareness of faults in complex industrial equipment, each client may not be able to observe every possible fault under all possible operating conditions.  In the domain adaptation problem, the partial domain adaptation (DA) scenario was defined by making the source domain have all classes of data and the target domain cover only a subset of those classes \citep{rombach2023controlled}. We extend this definition to FL scenarios where certain clients' datasets contain a subset of all classes. For consistency and simplification, such scenarios are also denoted as partial DA. The training datasets under scenarios 1 and 2 are defined using the partial DA setting with domain shift. Specifically for Scenario 3, we drastically varied the number of samples in each client's dataset to make the problem more difficult.
\begin{figure}[h!]
	\includegraphics[width=1\textwidth]{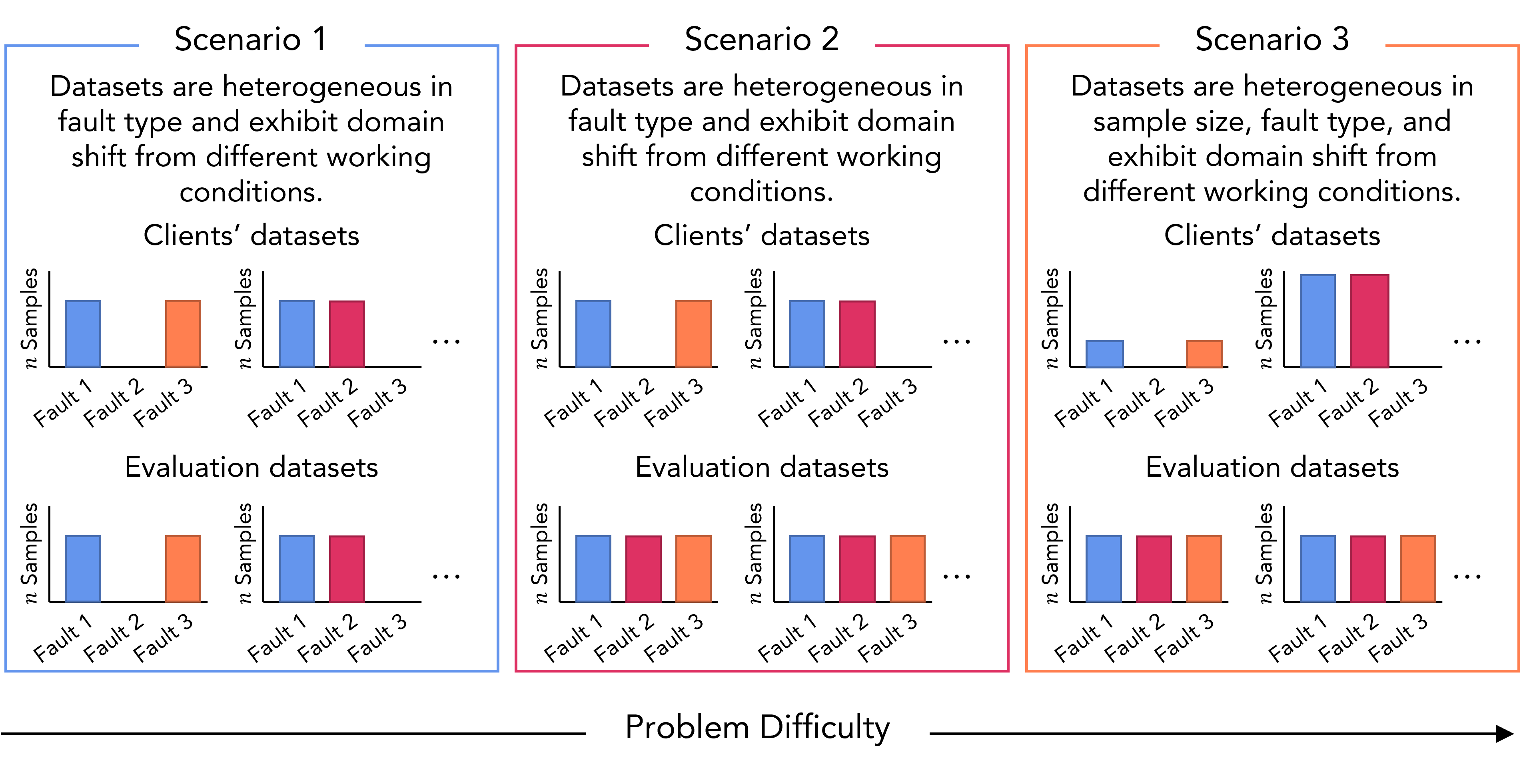}
	\centering
	\caption{An illustration of the three test conditions examined in this paper.}
	\label{fig:test_configurations}
	\centering
\end{figure}

Fig. \ref{fig:test_configurations} shows the three FL scenarios considered in this work. In scenario 1, client datasets are collected under different working conditions, and differences exist in each client's label space (partial DA setting). For example, some clients might only have healthy and inner race fault samples, while other clients might have healthy and outer race fault samples. Under Scenario 1, each client's evaluation dataset follows a similar distribution as the training dataset (the types of test data are identical to the training dataset). 

The training datasets in scenario 2 are identical to the settings in scenario 1. However, different from scenario 1, the test datasets are label-balanced datasets, which means clients need to collaborate to train a model that is capable of identifying unknown fault types.

Finally, for scenario 3, in addition to the feature distribution shift caused by different working conditions, the clients' datasets follow the partial DA setting with data quantity heterogeneity. The prediction difficulties are increased from scenario 1 to scenario 3. 

Three test scenarios are designed for each case study. We first divide each dataset into several subsets, where each subset contains data collected under the same operating condition. Then for each subset,  we further divide the subset dataset into several client local datasets. Table \ref{table:Exp_design} shows how the client training datasets are designed for each dataset case study.

For the CWRU dataset, we define six operating conditions, and $6\times 2 = 12$ client datasets are created, each containing data related to two particular health conditions. Under this setting, the clients need to work together to train a model that is capable of classifying three health conditions. For the PU dataset, $4\times 3 = 12$ client datasets are created. Clients 1,4,7 and 10 contain three types of training samples, while others only own two types of samples. The ISU dataset is also divided into 12 client datasets. 

In scenarios 1 and 2, all the training datasets are utilized to generate client training datasets, and in scenario 3, some clients are assigned a portion of the total data to create the client-wise quantified discrepancy. The detailed description of the dataset design is listed in the appendix tables \ref{table:Design_sc3train} and \ref{table:Design_test}.

\renewcommand{\arraystretch}{1.05}
\TableTwo
\renewcommand{\arraystretch}{1}

\subsection{Alternative FL Algorithms Used for Comparison}

The FedAvg \cite{mcmahan2017communication} and a cosine-similarity-based clustering algorithm are included as comparison models. In section \ref{sec:Performance Evaluation}, we compare the performance of the proposed FedSNGP against those two algorithms.

\begin{enumerate}[leftmargin=*,label=\alph*]
\item \textbf{Standard Locally-Trained Fault Diagnosis Models} To highlight the benefit of performing FL, we include the algorithm where each client updates their respective model parameters without FL. In the local training algorithm, each client trains the model for 250 epochs with a learning rate of 0.005.

\item \textbf{Federated Averaging} We use the federated averaging algorithm as a baseline model. As a widely recognized FL algorithm, FedAvg performs weighted averaging to aggregate all the client's parameters. The training procedure of FedAvg is given in Fig \ref{fig:Intro_FL} a.

\item \textbf{Federated Clustering Using Cosine Similarity} The cosine similarity has been widely utilized in clustering-based FL \citep{tian2022wscc,zhu2022aligning}. This study includs a clustering-based algorithm as a comparison model, denoted as FedCos. The FedCos algorithm differs from the FedSNGP in that it determines the clustering strategy by evaluating the cosine similarity between the parameters of local models instead of utilizing prediction uncertainty results. Given two local model parameters $\boldsymbol{\uptheta}^{i}$ and $\boldsymbol{\uptheta}^{j}$, the cosine similarity between the two models is defined as:
\begin{equation}
    Sim_{Cos}(\boldsymbol{\uptheta}^{i},\boldsymbol{\uptheta}^{j})=\frac{\boldsymbol{\uptheta}^{i}\cdot\boldsymbol{\uptheta}^{j}}{\lVert \boldsymbol{\uptheta}^{i} \rVert \lVert\boldsymbol{\uptheta}^{j} \rVert}
\end{equation}
The cosine similarity metric measures the angle between $\boldsymbol{\uptheta}^{i}$ and $\boldsymbol{\uptheta}^{j}$, a smaller angle indicates a higher similarity between  $\boldsymbol{\uptheta}^{i}$ and $\boldsymbol{\uptheta}^{j}$. Similar to the FedSNGP algorithm, each communication round of the FedCos algorithm is composed of four steps: 
\begin{enumerate}
\item Each client trains a local model individually and uploads the model to the server.
\item The server evaluates the cosine similarity between the  parameters of local models and generates a similarity matrix $M_{Sim}$.
\item The affinity propagation algorithm is applied to determine the clustering strategy and  generate an updated global model for each cluster by using Eqn. \ref{Eqn:Weighted_average}.
\item Each client downloads the updated global model from the server.
\end{enumerate}

\end{enumerate}
The key difference between FedAvg, FedCos, and FedSNGP is the server's model aggregation algorithm. Therefore, in this study, all the clients utilize the same SNGP model architecture for their models, and the processed features are a 1D array with a length of 512. The SNGP model is designed with three residual blocks, and each residual block contains one fully connected layer with  64 units. The models in this study are trained using FL algorithms for a total of 50 communication rounds. In each communication round, each client performs local model updates for a period of five epochs with a fixed learning rate of 0.005. 

\section{Results and Discussion}
\label{sec:Performance Evaluation}
\subsection{Evaluation Results}

Each of the 12 clients in the FL process undertakes their own diagnostic task,  resulting in a total of 12 tasks for each test scenario. To evaluate the performance of the algorithms, we begin by examining the average test accuracy across all 12 clients. The average test accuracies are summarized in Table \ref{table:Average_acc}, and the detailed results are listed in Appendix. Table \ref{table:All_results}

In scenario 1, each client's training and test data follow a similar distribution. The model training and test are conducted using samples that have the same types of bearing faults. The performance of FedAvg is worse than the other three methods, which is caused by the heterogeneity of the dataset. Still, all four algorithms achieved a test accuracy of more than 95\% for all three case studies.

\TableThree

Compared to scenario 1, scenarios 2 and 3 are more challenging, several clients have unbalanced training datasets, and the test dataset covers all the fault types. Clients need to acquire diagnosis knowledge from each other. Otherwise, it can not identify unknown fault types. Therefore, the performance of the local training algorithm experienced a significant decline. The following two subsections discuss the algorithms' performance in scenarios 2 and 3.

\subsubsection{Scenario 2 Case Studies}
\begin{figure}[H]
	\includegraphics[width=0.8\textwidth]{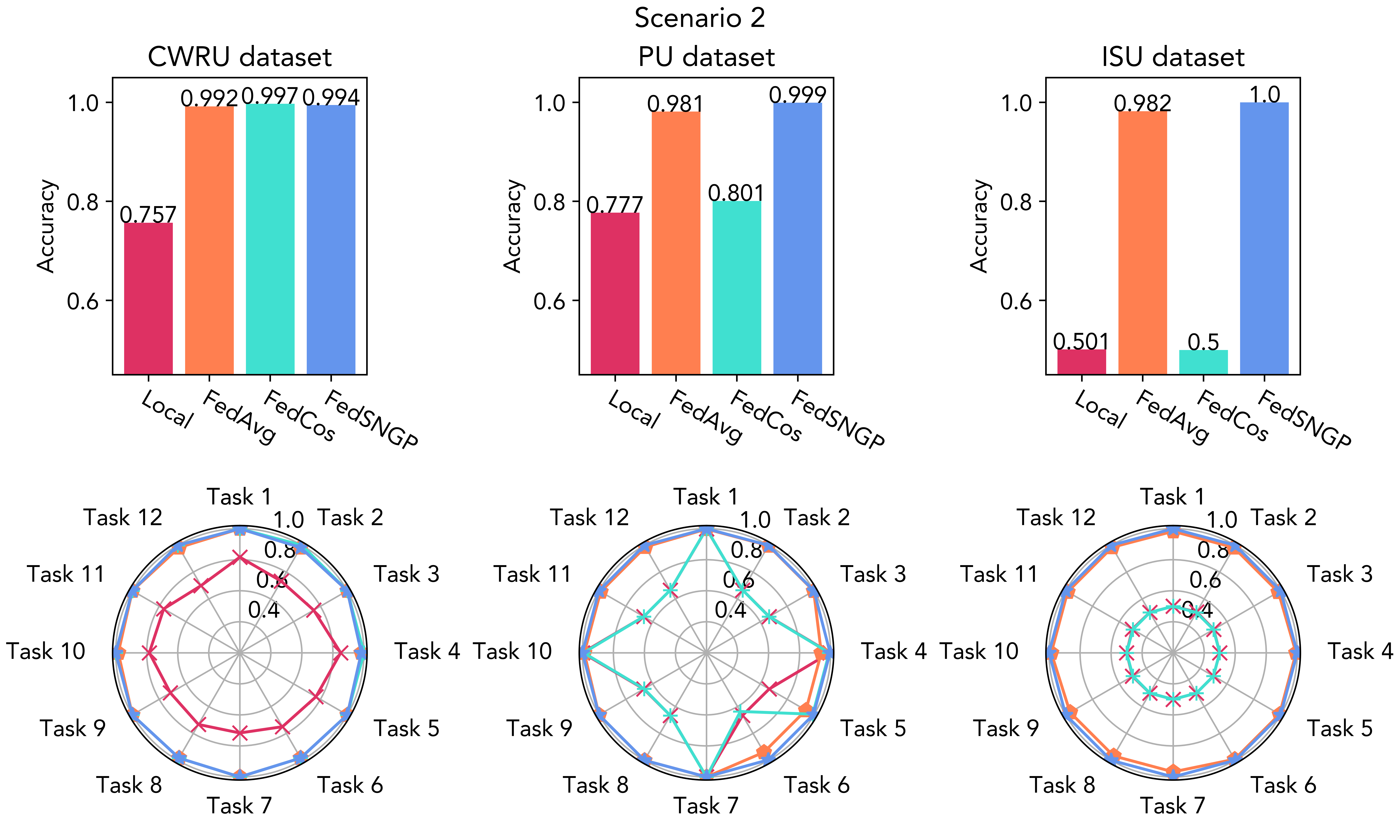}
	\centering
	\caption{The prediction results for scenario 2 case studies. The bar charts show the average test accuracy of each algorithm, and the radar chart shows the algorithms' prediction accuracy for each client.}
	\label{fig:results_SC2}
	\centering
\end{figure}
Figure \ref{fig:results_SC2} shows the prediction results for scenario 2. The top half of the figure summarizes the average test accuracy. And the bottom half of the figure shows the specific diagnosis accuracy for each client. 

As a benchmark bearing dataset, the fault pattern of the CWRU dataset is obvious, the change in working conditions does not lead to a significant domain shift, and the fault patterns can be identified easily. Considering that the feature distribution discrepancy is not severe in scenario 2, the FL algorithms successfully utilize the client's information, and both FedAvg, FedCos, and FedSNGP achieved a test accuracy of more than 95\%. As a comparison, the locally trained client models can not identify unknown fault types. Resulting in an accuracy of around $75\%$.

For the PU dataset case study,  the local datasets of clients 1, 4, 7, and 10 contain all the health conditions. That information is sufficient to train highly accurate diagnosis models. Therefore, the local training algorithm achieves a testing accuracy of around 100\%  for those four clients. On the other hand, the remaining eight client datasets were restricted to only one health and one faulty condition, preventing the client from training an accurate model independently without the knowledge of the missing fault types. As a result, the locally trained models for those eight clients were unable to effectively diagnose fault types that were not present in their respective training datasets. As a benchmark FL algorithm, FedAvg aggregates all the client's information by performing weighted averages to local models' parameters. Most of the models trained by FedAvg achieve an accuracy of more than 95\%, and clients 4,5,6 yield a 94.00 \% accuracy. This decline in performance is attributed to the domain discrepancy among clients, primarily due to the variance in working conditions. The average accuracy of FedCos is 80.1\%, which is slightly higher than the one of the local training method (77.7\%). Based on the radar chart analysis, the performance of FedCos models was inferior to that of FedAvg models. The models trained by FedCos yield 100\% accuracy on clients 1 and 7, and for other clients, the model's accuracy is around $2/3$. This is mainly because FedCos is not able to accurately cluster the clients. The FedSNGP models achieved almost 100\% accuracy for all the clients. Finally, the ISU dataset case study shows that the proposed FedSNGP outperforms the other three methods.

\subsubsection{Scenario 3 Case Studies}

\begin{figure}[H]
	\includegraphics[width=0.8\textwidth]{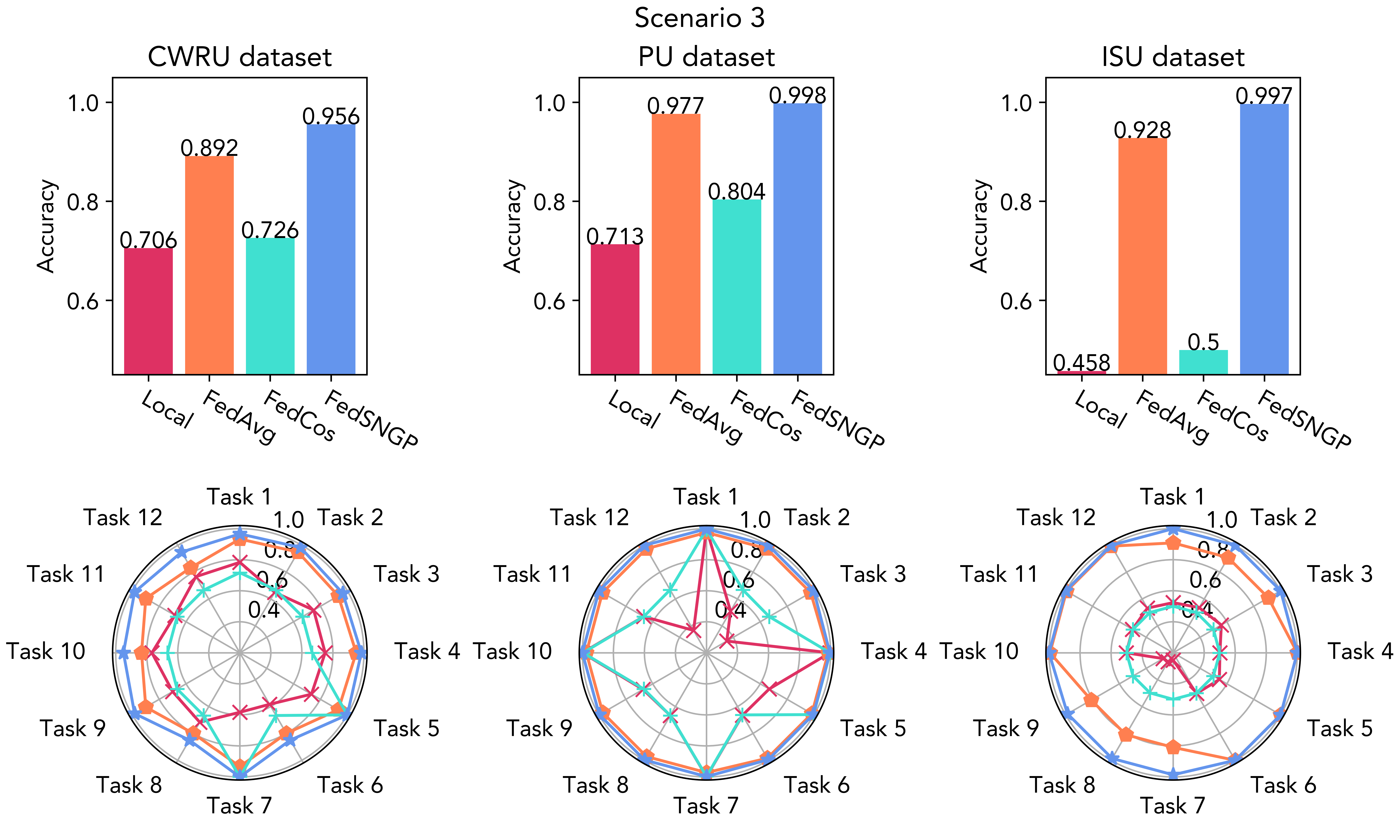}
	\centering
	\caption{The prediction results for scenario 3 case studies. The bar charts show the average test accuracy of each algorithm, and the radar chart shows the algorithms' prediction accuracy for each client.}
	\label{fig:result_SC3}
	\centering
\end{figure}
In addition to domain shifts and partial DA setting, scenario 3 also considers differences in the number of training samples. In this scenario, some clients may possess a larger number of samples, while others may have only a limited number of samples.  Previous studies have indicated that unbalanced training sample quantities can have a  significant impact on the performance of the FedAvg algorithm \citep{zhu2022aligning}. The evaluation results for scenario 3 case studies are summarized in Fig  \ref{fig:results_SC2}. The feature distribution and sample quantity discrepancy make diagnosis tasks more challenging than scenario 2. As a result of the varying numbers of training samples held by each client, the training dataset exhibits more pronounced Non-IID properties, resulting in the performance decline of FedAvg and FedCos models. However, the introduction of training sample number discrepancy does not affect the performance of the proposed FedSNGP. In all three case studies within scenario 3, models trained using FedSNGP outperformed those trained using other methods, achieving average prediction accuracies higher than 95\%.

\subsection{Analysis of the Clustering Strategy}

Unlike FedAvg, which creates a single global model for all the clients, clustering-based FL algorithms assign clients to multiple clusters. Then, the clients in each cluster are federated together to train a global model. Several cosine similarity-based clustering FL algorithms have been developed \citep{tian2022wscc,li2021federated}. This study considers both domain shift and partial DA setup. It is surprising that the FedCos algorithm does not show any improvement compared to FedAvg. In this section,  we use the PU case study (scenario 2) to compare the clustering strategy of FedSNGP and FedCos and analyze why FedCos performs worse than FedSNGP.

\begin{figure}[h!]
	\includegraphics[width=0.8\textwidth]{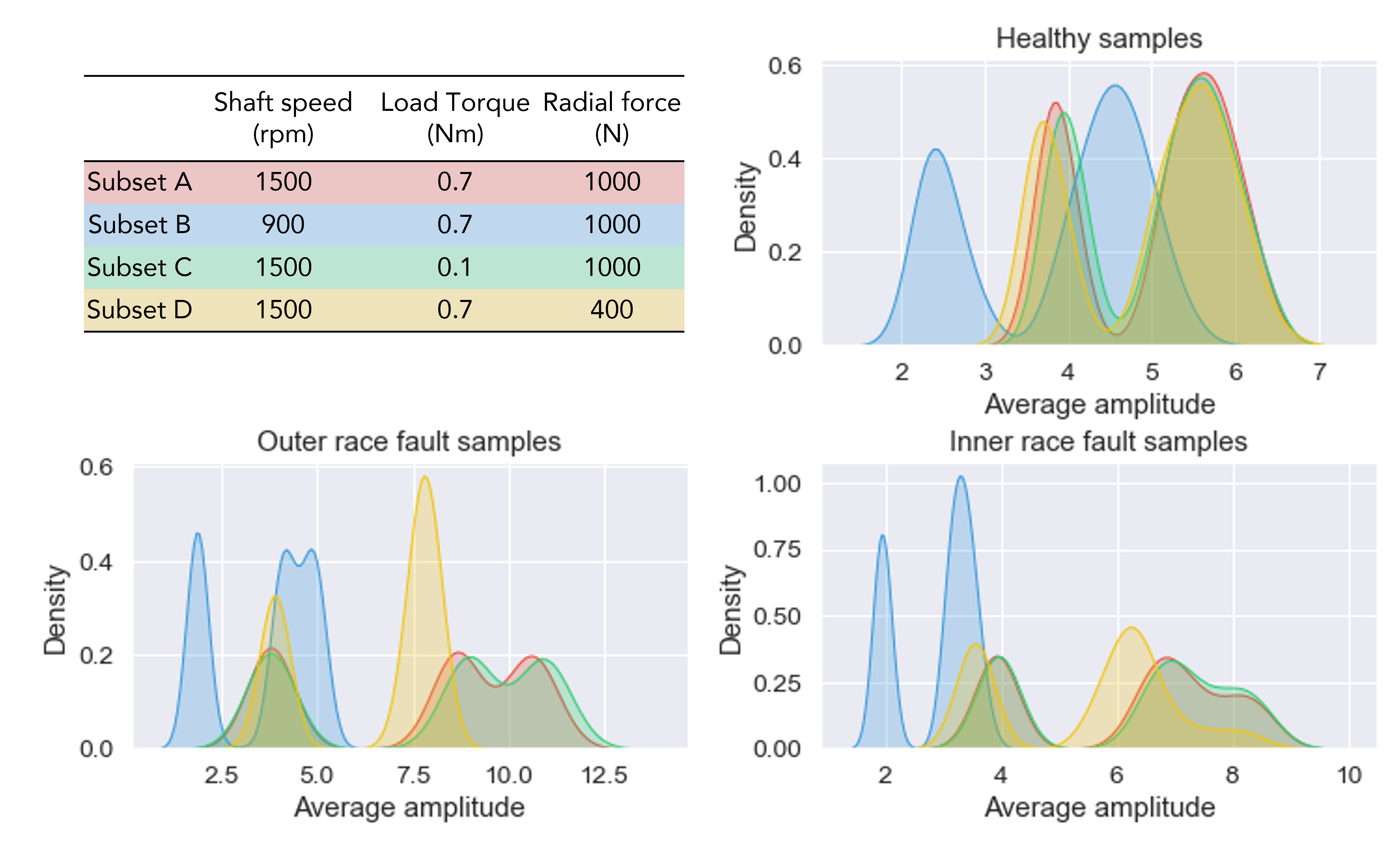}
	\centering
	\caption{Kernel density estimations of datasets by calculating the average amplitude of input features (sorted by different labels)}
	\label{fig:KDA_PU}
	\centering
\end{figure}

The PU dataset was collected under four working conditions, with each working condition forming a subset. Fig \ref{fig:KDA_PU} visualizes the distribution of the average amplitude of the samples. It has been acknowledged that the shaft speed affects the bearing fault characteristic frequencies, and the change of shaft speed affects the vibration amplitude a lot. The subset B data (marked by blue shade)  was collected under 900 rpm, which is significantly different from the other three subsets. The kernel density estimation results also infer that the feature distribution of subset B is significantly different from the other three subsets. For the remaining three subsets, their healthy samples seem to follow a similar distribution. The feature distributions of outer race and inner race fault samples suggest that data in subset D  differs slightly from the other two subsets. 
Based on the kernel density estimation analysis, it appears that subsets A and C have similar data distributions, while subset D has some differences from subsets A and C. Subset B, on the other hand, is notably  different from the other subsets. A desired clustering strategy needs to make sure that: a) the clients in the same subset (which means the client datasets were collected under the same working condition) should be grouped together; and b) if two subset features share similar distribution, then those two subsets should be grouped together. Following this assumption, the optimal clustering result for the PU case study should result in three clusters: {client 1, client 2, client 3, client 7, client 8, client 9},{client 3, client 4, client 5}, and {client 10, client 11, client 12}.

\begin{figure}[h!]
	\includegraphics[width=0.75\textwidth]{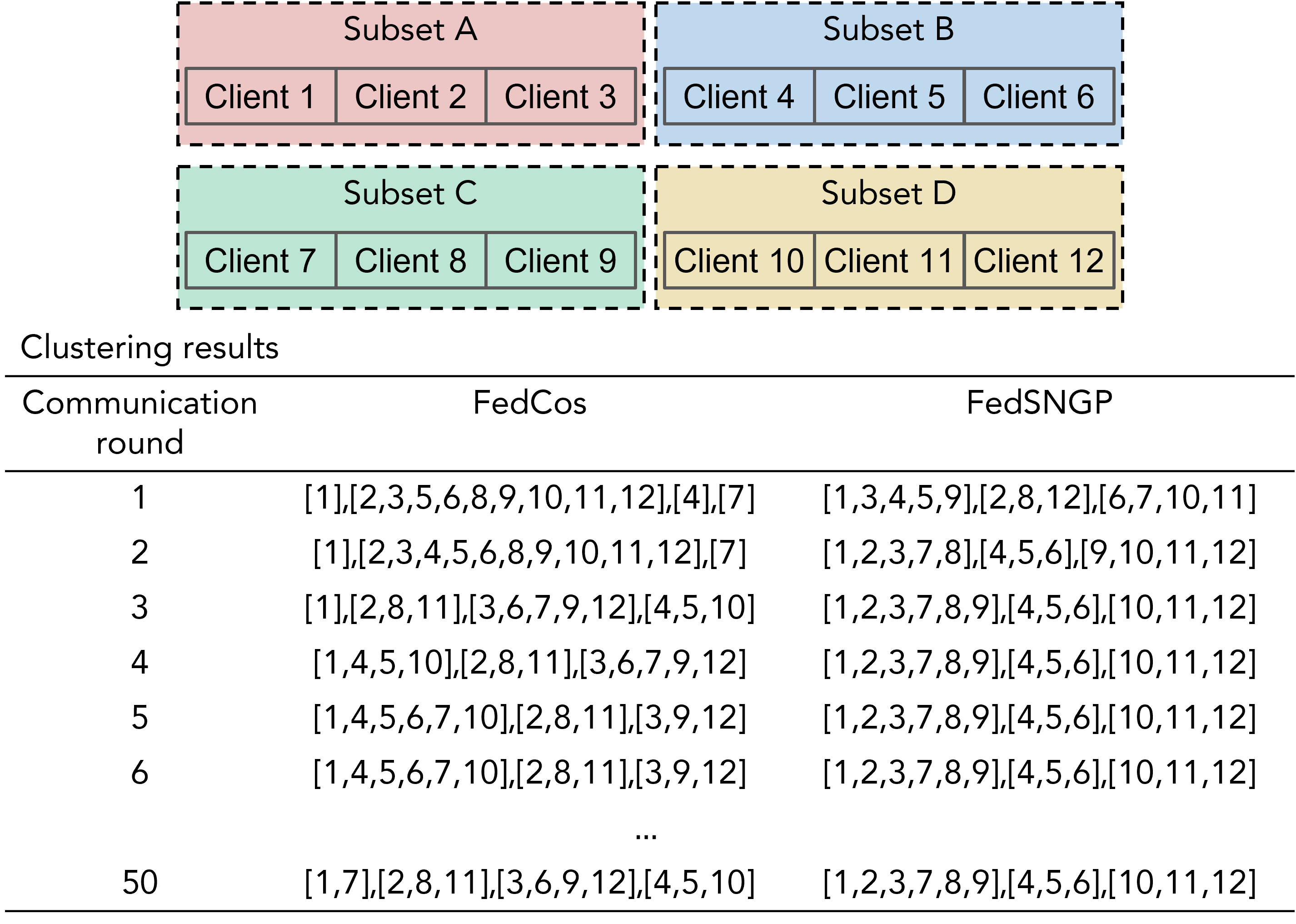}
	\centering
	\caption{Clustering results provided by FedCos and FedSNGP}
	\label{fig:Clustering Results}
	\centering
\end{figure}

Fig. \ref{fig:Clustering Results}. shows the dynamic changes in clustering results provided by FedCos and FedSNGP in the different communication rounds.  In the first communication round, since the models are still in the early training stage, both FedCos and FedSNGP can not group clients correctly. As the number of communication rounds increased, each client model started to capture the distribution of its training dataset. FedSNGP assigns clients into three clusters. The subsets A and C form one cluster, while B and subset D clients form two clusters. On the contrary, the FedCos fails to group clients from the same subset into the same cluster. The incorrect clustering strategy employed by FedCos resulted in a decrease in the model's performance.

\begin{figure}[h!]
	\includegraphics[width=1\textwidth]{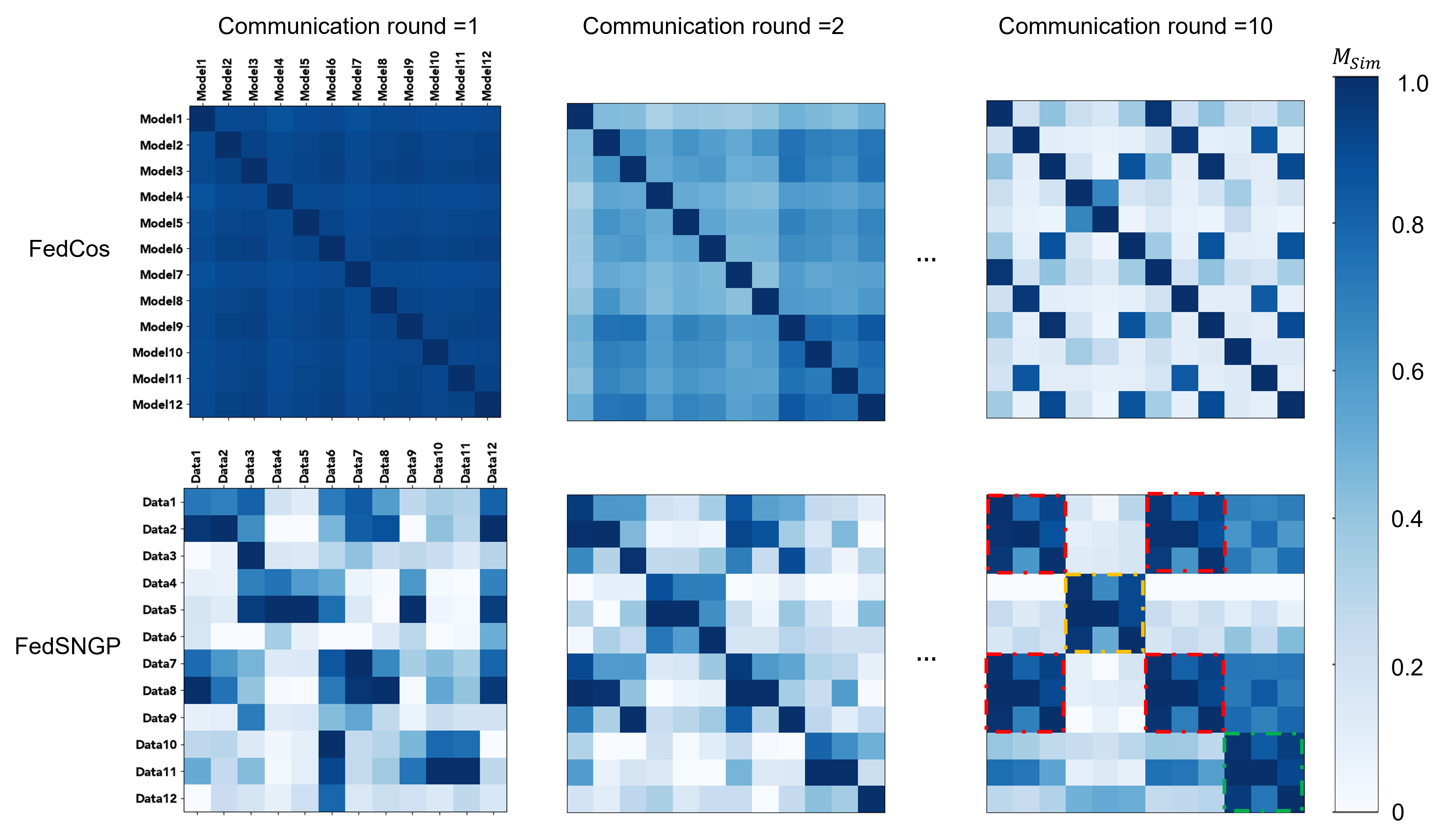}
	\centering
	\caption{Compare the similarity matrix generated by FedCos and FedSNGP at communication round 1, 2, and 10}
	\label{fig:Clustering}
	\centering
\end{figure}

Fig \ref{fig:Clustering}  shows the dynamic changes of the similarity matrix provided by FedCos and FedSNGP in each communication round. Each entry shows the calculated similarities between the two models. The darker blue indicates a higher similarity between the two clients. It should be noted that while FedCos and FedSNGP utilize the same model architecture, they use different metrics to infer data distribution similarity. FedCos estimates the similarity by comparing the model parameters. In the initial communication round, since each client model is initialized with the same parameters and has undergone only five local training epochs, the cosine similarity matrix between each pair of clients exhibits large values. After the second communication round, it is still difficult to determine if any two local models are significantly different. After the communication round 10, the results start to indicate high similarities between certain client models. However, due to the severe distribution discrepancy among clients, some clients are optimized in different directions. More importantly, within each communication round, the server clusters clients based on the model parameters, and the model aggregation process affects the updated model parameters. In case a client is assigned to an inappropriate cluster, it becomes challenging for the algorithm to rectify the mistake and assign the client to the correct cluster. As a result, the $M_{Sim}$ generated by FedCos does not necessarily reflect the dataset similarity, and the algorithm does not assign clients of the same subset to the same cluster.  For example, we expect the models of clients 1, 2, 3, 7, 8, and 9 to show a higher similarity because those three clients collect data under the same working condition (subset A). However, according to the entries of the first row, the normalized cosine similarity between clients 1 and 7 is significantly larger than other entries, while the cosine similarity between 1 and 2 is relatively low. Finally, only clients 1 and 7 are clustered together.

Unlike FedCos, which uses cosine similarity to cluster clients, the FedSNGP uses model prediction uncertainty as a driving force for clustering.  FedSNGP generates a similarity matrix ($M_{Sim}$)  that is non-symmetric, indicating that the similarity between two datasets is not necessarily the same in both directions. By considering prediction uncertainty, the FedSNGP is better equipped to estimate dataset similarity, making it more effective than FedCos in clustering clients for FL. 

While after the first communication, the FedCos indicates that all the models are similar (with cosine similarity being more than 0.8 for each pair of clients), FedSNGP starts to indicate that certain client datasets are significantly different from others. For example, the second row indicates that dataset 2 is significantly different from dataset 4 and 5 (the similarity values are around 0). It is worth noting that a single client's estimation results can be unreliable. For instance, if we consider the perspective of client 3  (the third row), its data distribution is significantly different from that of clients 1 and 2. This is due to the fact that the training dataset of client 3 only covers healthy and outer race fault samples, unlike client 1,which has all the classes of training datasets. Consequently, the similarity estimations from client 3 may not be as dependable as those from client 1. While both the first row and second row indicate that clients 1,2,3,6,7 and 12 have similar data distributions,  the third row suggests that the dataset of client 3 differs significantly from all the other clients. And affinity propagation clustering considers all clients' information and creates client clusters in an adaptive manner. Notably, after  round 10 of communication, the FedSNGP algorithm formed three clusters: the first cluster included clients 1,2,3,5,6, and 7, the second cluster grouped clients 3,4,5, and the third cluster included clients 10,11,12.  

Figures \ref{fig:Clustering Results} and \ref{fig:Clustering} illustrate how clients are clustered using FedCos and FedSNGP, respectively. Fig. \ref{fig:Clustering Results} analyzes the training datasets by examining the average amplitude. To validate the accuracy of the clustering strategy, we conducted the following test.

(a) We divide the whole dataset into four subsets based on the working conditions. We then use each subset to train an individual SNGP model. The total number of training epochs is 20. After every five epochs, we record the model parameters. Finally, we use the principal component analysis algorithm to map the parameters of each model onto a 2D plot.

(b) In the context of the PU dataset (scenario 2 setting), we utilized 12 client datasets to train 12 models. The number of training epochs is 20. Again, model parameters are recorded at every five epochs. Finally, we use principal component analysis to map the parameters of the 12 models onto a 2D plot.
\begin{figure}[h!]
	\includegraphics[width=1\textwidth]{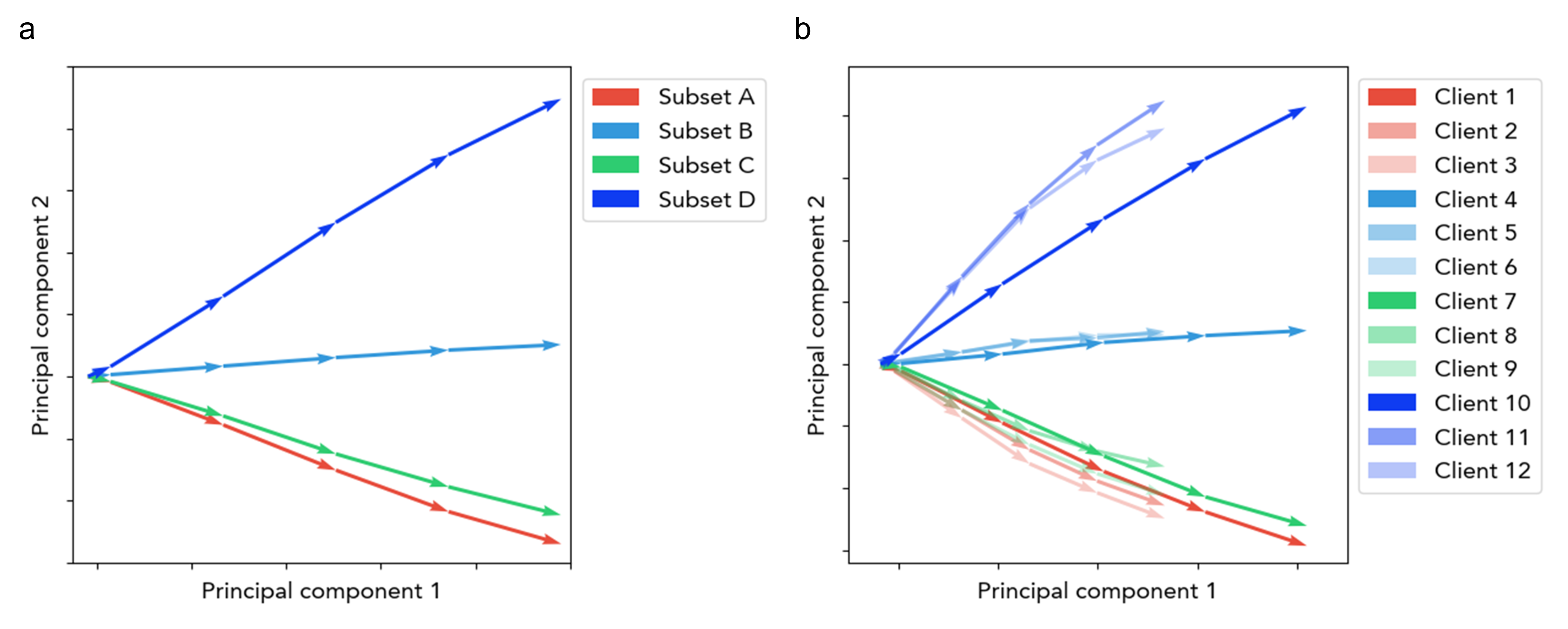}
	\centering
	\caption{Change of model parameters. The model parameters are mapped into 2D plots, and the arrow connects each adjacent model point to show the optimization direction of each model.}
	\label{fig:Optimization_direction}
	\centering
\end{figure}
As shown in Fig\ref{fig:Optimization_direction}a, the shift of model parameters indicates that models trained on subset A or C are being optimized in a comparable direction, suggesting that the data distribution of subset A is similar to subset C. According to our kernel density feature distribution analysis, subsets B and D should be trained independently without utilizing information from other subsets. This is further supported by Figure \ref{fig:Optimization_direction}b, which demonstrates that the client models generally form three clusters. However, for clients with imbalanced training datasets, such as clients 11 and 12, their optimization directions exhibit slight divergence from the desired direction, especially during the initial stages. This divergence may potentially mislead the cosine-similarity-based FL approach. Therefore, these findings suggest that at the early stages of local model training, the similarity of model parameters may not necessarily reflect the similarity or discrepancy of the data. 

\subsection{Handling Out-of-Cluster-Distribution Samples}

One limitation of clustering-based FL is that the trained model is only applicable to the data distribution within the same cluster. In the context of bearing diagnosis, if the testing data falls outside the distribution of the client's training dataset, the client may generate inaccurate predictions which can have a substantial impact on availability and maintenance costs. This limitation poses a challenge to the practical application  of the diagnosis algorithm in industrial settings. However, unlike conventional clustering algorithms, the FedSNGP incorporates prediction uncertainty into the FL framework, which can help clients in avoiding overconfident predictions for out-of-distribution samples. This property of uncertainty quantification could potentially enhance the reliability and practicality of the FedSNGP algorithm for industrial applications.

In this section, we show how FedSNGP deals with out-of-distribution samples. If an SNGP model is tested with out-of-distribution samples, the prediction uncertainty increases significantly higher compared to the uncertainty towards the training dataset. To mitigate this issue, a prediction uncertainty threshold is defined based on the model's training dataset predicted variance. In this study, we set the threshold to ten times the training dataset predicted variance. If the predicted variance for a test dataset exceeds this threshold, it suggests that the prediction is unreliable, and clients seek help from other clients.  

\begin{figure}[h!]
	\includegraphics[width=0.5\textwidth]{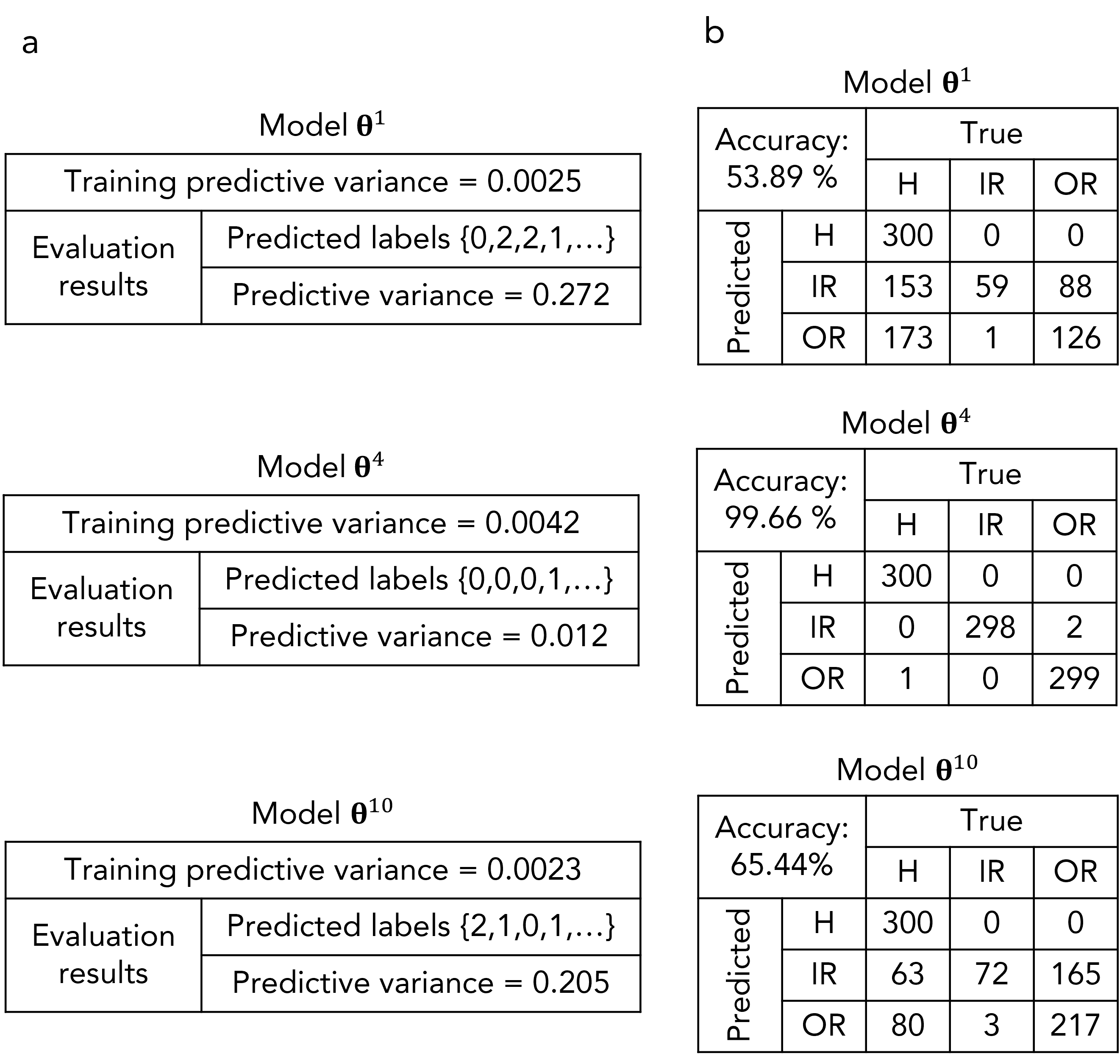}
	\centering
	\caption{An example of predicted variance-guided model selection: a. the model's predicted variance results. b. the confusion matrix of prediction results }
	\label{fig:OOD_detecion}
	\centering
\end{figure}

To do this, the client downloads other client clusters' models from the server and tests them with the test dataset. The client selects the model that yields the least predicted variance. Fig \ref{fig:OOD_detecion} provides a numerical example using client 1 in the PU dataset case study (scenario 2). The client first evaluates the test dataset using model $\boldsymbol{\uptheta}^{1}$, with a test predicted variance of $0.272$, which is significantly larger than its predicted variance towards its training dataset. The client then seeks help from the server.

On the server side, it is known that clients form three clusters. The server shares the other two clusters' models ($\boldsymbol{\uptheta}^{4}$ and $\boldsymbol{\uptheta}^{10}$) with client 1. The model $\boldsymbol{\uptheta}^{4}$ yields the least predicted variance, and that value is within the threshold. Therefore, model $\boldsymbol{\uptheta}^{4}$'s result is more reliable. The confusion matrix listed in Fig \ref{fig:OOD_detecion} b demonstrates that $\boldsymbol{\uptheta}^{4}$ yields the highest accuracy. The predicted variance-guided model selection strategy can fully utilize the information provided by all the clients.

\section{Conclusion}
\label{sec:Conclusion}
This study proposes an uncertainty-based dynamic clustering federated algorithm, called FedSNGP, for fault diagnosis on fleets of system or component units, considering partial DA setting and feature shift. The proposed FedSNGP clusters clients based on inferred dataset similarity and clients within the same cluster federate to train a model applicable to its own data distribution. FedSNGP has two desirable features: (1) it uses predicted variance results instead of model parameter similarity to infer dataset similarity and does not require defining the number of clusters, and (2) it detects out-of-distribution samples and prevents overconfident predictions.

Experimental results using three datasets and three FL scenarios demonstrate that FedSNGP outperforms FedAvg and FedCos, achieving over 99\% accuracy for all clients.  The proposed model is applicable to severe partial DA scenarios, and by analyzing the distribution of client datasets and model parameter changes during training, it accurately defines clusters and yields the highest prediction accuracy.

However, there are still limitations that need future research. Specifically, estimating data similarity requires clients to download multiple models from the server, resulting in increased computational resources and training time compared to FedAvg. A potential strategy to mitigate this limitation is to reduce data transmission volume between the server and clients by encoding model parameters. Additionally, our strategy requires each client to have access to the other clients' models, which may lead to privacy concerns. One way to mitigate this issue is to shuffle the order of client models and hide the models' client IDs when a client downloads other models. This makes it more challenging for each client to reconstruct other clients' training datasets.

\section*{Acknowledgements}

This work was supported in part by the U.S. National Science Foundation under Grant IIP-1919265. Any opinions, findings, or conclusions in this paper are those of the authors and do not necessarily reflect the sponsor’s views.

\bibliographystyle{unsrtnat}
\bibliography{References}  
\section*{Code Availability}
The dataset and related codes are provided at \url{https://github.com/SalieriLu/FL_fault_diagnosis}

\section*{Appendix}

\subsection*{Design of the training datasets}
To create data sample heterogeneity, in scenario 3, some clients only use a portion of the training dataset to train the model. The design of the training dataset for scenario 3 is listed as follows:
\Appendtrainingsc
\subsection*{Design of the test datasets}
The design of test datasets is summarized in the following table:
\Appendtestdataset
Note that scenario 2 and 3 uses the same test dataset setting.

\subsection*{Results for each of the three test conditions}
\Appendresults

\end{document}